\documentclass{article}
 \usepackage[preprint]{neurips_2026}

\usepackage{pifont}

\usepackage[utf8]{inputenc} 
\usepackage[T1]{fontenc}    
\usepackage{hyperref}       
\usepackage{url}            
\usepackage{booktabs}       
\usepackage{amsfonts}       
\usepackage{nicefrac}       
\usepackage{microtype}      
\usepackage{xcolor}         
\usepackage{amsmath}
\usepackage{graphicx}
\usepackage{multirow}
\usepackage{wrapfig}
\usepackage[table]{xcolor}
\usepackage{placeins}

\title{HEART: Hyperspherical Embedding Alignment via Kent-Representation Traversal in Diffusion Models}

\author{%
 Arani Roy$^*$ \\
  Purdue University\\
  \texttt{roy173@purdue.edu} \\
  \And
   Shristi Das Biswas$^*$ \\
  Purdue University\\
  \texttt{sdasbisw@purdue.edu} \\
    \And
   Kaushik Roy \\
  Purdue University\\
  \texttt{kaushik@purdue.edu} \\
}

\begin{document}

\maketitle

\begin{abstract}
Text-to-image diffusion models can generate visually stunning images, yet, controlling what appears and how it appears - remains surprisingly difficult, especially when operating solely within the constraints of the text-conditioning space. For example, changing a subject or adjusting an attribute often leads to unintended side effects, such as altered backgrounds or distorted details. This is because most existing text-based control methods treat the embedding space as Euclidean and apply simple linear transformations, which do not reflect how semantic concepts are actually organized. In this work, we take a step back and ask: what is the true geometry of these embeddings? We find that text encoder representations lie on a hypersphere, where concepts are not linear directions but structured, anisotropic distributions better captured by Kent distributions. Building on this insight, we propose \textbf{HEART}, a training-free framework that performs Kent-aware geodesic transformations directly on the hypersphere. By respecting the underlying geometry, HEART enables intuitive and precise edits, such as consistent subject replacement and fine-grained attribute control, while preserving the original scene. Importantly, HEART requires no finetuning, inversion, or optimization, and generalizes across diffusion model architectures. Our results show that a simple shift in perspective, from linear to spherical, can unlock fast, and controllable image generation.
\end{abstract}

\section{Introduction}
Text-to-image (T2I) diffusion models have enabled remarkable progress in controllable image generation, allowing users to specify desired content through natural language prompts (\cite{rombach2022high,podell2023sdxl,esser2024scaling}). 
However, beyond generation, modern applications also require \emph{precise and nuanced control} over both subjects and their attributes, for example, replacing a subject \textcolor{red}{``cat''} with another subject \textcolor{cyan}{``dog''} (\emph{subject control}) or transitioning an attribute \textcolor{red}{``young''} face to \textcolor{cyan}{``old''} one (\emph{attribute control}). Achieving this level of editing granularity remains a formidable challenge, as it requires carefully isolating and manipulating specific semantic components while preserving the broader scene structure.

A growing line of work seeks to address this problem through various control paradigms, ranging from cross-attention manipulation (\cite{hertz2022prompt}) and semantic guidance (\cite{brack2023sega}) to parameter-efficient fine-tuning via LoRA adapters (\cite{gandikota2024concept, chiu2026text, sridhar2024prompt}). While these approaches have shown promising results, achieving precise and localized semantic control often requires additional computational mechanisms to mitigate semantic entanglement and background drift. In particular, consistent subject-level editing frequently relies on expensive inversion pipelines (\cite{hertz2022prompt}), per-concept optimization (\cite{baumann2025continuous}), attention-map manipulation(\cite{hertz2022prompt}), or specialized adapter training(\cite{chiu2026text, gandikota2024concept}), limiting scalability and generalization across concepts and diffusion backbones. Moreover, most prior approaches operate under a common assumption: concepts correspond to linear directions in embedding space. Under this view, semantic edits are performed through vector arithmetic, adding or interpolating along a direction associated with a concept.

In this work, we revisit the geometry of text encoder representations in diffusion models and uncover a fundamentally different structure. Prior research like~\cite{levi2024double} has probed into analysis of CLIP-based encoder geometry~(\cite{radford2021learning}) and has interestingly revealed that the raw embedding space assumes a ``thin-shell'' ellipsoid geometry (thin-shell referring to the embeddings' distance from the origin being much larger than the variation in their magnitudes). Also, (\cite{wang2020understanding}) state that the contrastive learning objective used during pre-training forces these output representations of the encoders to lie on a spherical surface to maximize alignment and uniformity. Motivated by these ideas coming together, we empirically study text embeddings across a wide range of diffusion architectures, including both U-Net (SD1.5 \citep{rombach2022high}, SDXL \citep{podell2023sdxl}) and DiT-based models (SD3.5-medium \citep{esser2024scaling}, FLUX \citep{flux2024}). We confirm that across all settings, we find a consistent pattern: semantic information is primarily encoded in the direction of embeddings on a hypersphere, rather than in their radial magnitude (Section \ref{sec:hypersphere}).
However, this directional view alone is insufficient. Prior work commonly assumes that concepts correspond to single linear directions or subspaces in embedding space. We show that this assumption is overly restrictive. Instead, concepts form anisotropic distributions on the hypersphere (Section~2.2), reflecting structured variation along multiple semantic axes. Empirically, we find that these distributions are best modeled by the Kent distribution (\cite{kent1978some, kasarapu2015modelling}), a hyperspherical analogue of an elliptical Gaussian distribution, which consistently outperforms isotropic alternatives across multiple diffusion architectures (Section \ref{sec:concept_geometry}). This geometric insight helps explain why many existing text-based editing methods can produce semantic drift and structural distortions. By performing linear operations in a curved hyperspherical space, these approaches may push embeddings away from the semantic manifold into "off-manifold" lower-density regions of representation space, resulting in unstable or entangled edits during generation.

Beyond this geometric structure, we further identify a phenomenon of token-level contamination in text encoder representations (Section \ref{sec:token_contamination}): a mechanism wherein concept-specific information propagates across the token sequence through attention. In causal encoders (CLIP), this manifests as directional propagation from early tokens to downstream positions, whereas in bidirectional encoders (T5), concept information is distributed more uniformly across both upstream and downstream tokens. This provides a principled explanation for background drift. Since scene-level semantics are often anchored in these downstream or upstream tokens, its modifications lead to unintended changes in the overall image. Achieving precise control thus requires a dual strategy: geometry-aware transformations to respect the manifold’s curvature and position-aware editing to isolate a concept’s influence without disturbing the global scene.

Building on these insights, we propose \textbf{HEART} (Hyperspherical Embedding Alignment via Kent-Representation Traversal), a geometry-driven framework for controllable T2I generation. HEART operates directly within the text embedding space by performing geodesic traversal on the hyperspherical manifold, guided by the directional statistics of the Kent distribution. For \textbf{subject control (HEART-S)} (e.g. \textcolor{red}{cat} $\rightarrow$ \textcolor{cyan}{dog}), we introduce a position-aware geodesic rotation at the concept token that replaces the source subject while strictly preserving the semantic integrity of the downstream context. For \textbf{attribute control (HEART-A)}, we identify the attribute direction (e.g. \textcolor{red}{young} $\rightarrow$ \textcolor{cyan}{old}) within the Kent distribution of the concept and perform a localized geodesic traversal, enabling fine-grained edits that are disentangled from the underlying concept identity.

Notably, HEART is \emph{training-free} and operates entirely at inference time, requiring no fine-tuning, per-concept optimization, or inversion.  Furthermore, the framework is architecture-agnostic, generalizing across diverse diffusion backbones. We summarize our contributions as follows: \textbf{1) Geometric Characterization of Text Embedding Space.} We provide a rigorous analysis of the text embedding space in diffusion models, unveiling a hyperspherical geometry where concepts are represented by anisotropic Kent distributions. We also examine contamination in the token distribution space, which, along with the denoising step at which modified text embeddings are introduced, explains the background drift observed in T2I models. \textbf{2) HEART Framework.} We propose HEART, a training-free, geometry-consistent framework that enables subject and attribute control via geodesic traversal. By operating on the manifold's natural curvature, HEART achieves precise modifications without the computational overhead of fine-tuning or inversion. \textbf{3) Comprehensive Empirical Validation.} We validate HEART across four diffusion architectures: SD1.5, SDXL, SD3.5, and FLUX, benchmarking against state-of-the-art baselines. We demonstrate that HEART establishes a new frontier for real-time, training-free T2I control that scales across diverse concepts and model families.

\section{Revealing True Nature of Text Encoder Geometry}
\label{sec:geo}
In this section, we characterize the geometry of text encoder representations used in different diffusion models. First, in Section~\ref{sec:hypersphere}, we demonstrate that semantic relationships are governed by the \emph{angular structure} of embeddings, while magnitude does not change the semantic meaning itself. Next, in Section~\ref{sec:concept_geometry}, we demonstrate that individual concepts (e.g., \texttt{cat}) are not represented as single directions, but as anisotropic Kent clusters on the hypersphere. Finally, in Section~\ref{sec:token_contamination}, we analyze how semantic information propagates across token representations in diffusion conditioning, including end-of-text and padding tokens, revealing a form of causal contamination in token embeddings. Together, these findings provide a unified geometric perspective on text embeddings in diffusion models, which we leverage for precise and structured semantic control.
\begin{figure*}[t]
\centering
\begin{minipage}[t]{0.52\textwidth}
\vspace{0pt}
\centering
\small
\setlength{\tabcolsep}{3pt}
\renewcommand{\arraystretch}{1.08}

\resizebox{\linewidth}{!}{
\begin{tabular}{l|l|l|l}
\toprule
Model & Query & Linear NN & Angular NN \\
\midrule
SDXL-G & cat 
& \texttt{emoji, teamsisd, kerswednesday} 
& \texttt{cats, cat, kitten} \\

SDXL-G & car 
& \texttt{btsbbmas, swachhb, nswpol} 
& \texttt{cars, car, vehicle} \\

Flux-L & cat 
& \texttt{B, [id=197], ctrl-token} 
& \texttt{cats, cat, dog} \\

Flux-L & car 
& \texttt{A, J, B} 
& \texttt{cars, car, vehicle} \\
\bottomrule
\end{tabular}
}
\footnotesize
\textbf{Nearest neighbors (NN) in token space.} We obtain semantically unrelated tokens by traversing across linear space, while traversal on the spherical surface using angular (cosine) similarity yields semantically coherent concepts.

\end{minipage}
\hfill
\begin{minipage}[t]{0.44\textwidth}
\vspace{0pt}
\centering
\includegraphics[width=\linewidth]{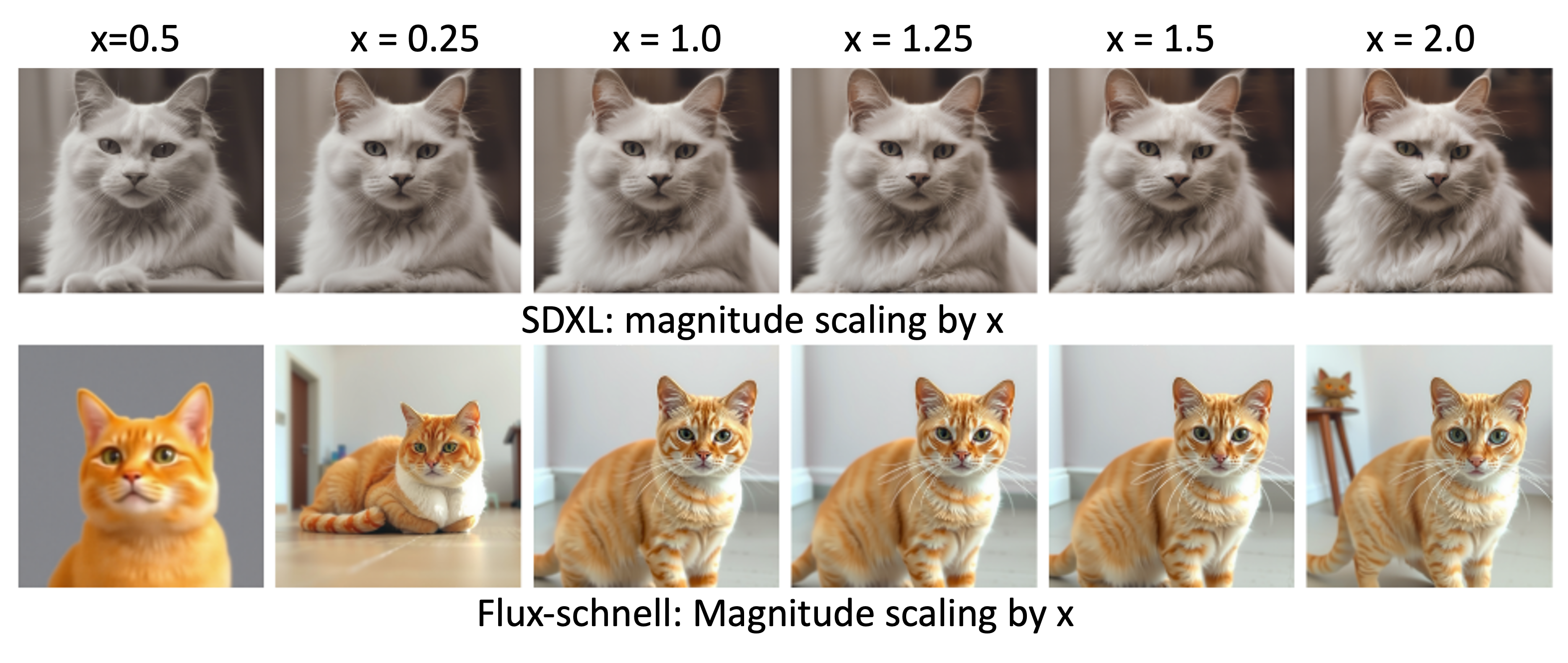}
\footnotesize
\textbf{Magnitude scaling.}
Magnitude scaling of embeddings does not change semantic output across models. (Prompt:``A cat'')
\end{minipage}
\vspace{4pt}
\caption{
Directional semantics in diffusion text embeddings. Left: Across SDXL OpenCLIP-G and Flux CLIP-L models, angular similarity retrieves meaningful neighbors while linear distance does not. Right: Scaling embedding magnitude does not affect generation, suggesting that magnitude plays a limited role in semantic content.
}
\label{fig:mag_and_nn}
\end{figure*}

\subsection{Hyperspherical Structure of Text Embeddings}
\label{sec:hypersphere}
Prior work like~\cite{levi2024double} show that CLIP embeddings lie on thin, ellipsoidal shells, where ``thin'' means that the variation in embedding magnitude is small. At the same time, (\citet{wang2020understanding}) demonstrate that contrastive learning objectives, used in CLIP-based text encoders, encourage representations to align based on angular similarity. Additionally,~\cite{feng2026beyond, levi2024double} find that variations in magnitude do not encode semantic differences; instead, they are associated with factors such as concept rarity or confidence. 
These works primarily study the geometry of pooled CLIP embeddings(\cite{radford2021learning}). However, T2I diffusion models condition not only on pooled representations, but also on full token sequences. It therefore remains unclear whether the directional and hyperspherical properties observed in pooled embeddings also extend to the token-level representations used during diffusion conditioning. To investigate this, we analyze the text-conditioning embeddings used in SD1.5 (CLIP-L), SDXL (OpenCLIP-G), SD3.5 (CLIP-L/G, T5-XXL), and FLUX (CLIP-L, T5-XXL). We observe that embeddings across all CLIP-based encoders concentrate on a remarkably thin spherical shell (Appendix. ~\ref{geo:textspace}.1), indicating low variation in embedding norms. In contrast, T5-based embeddings exhibit higher norm variability, suggesting a weaker thin-shell structure. This shows that, at least, for CLIP encoders, the embedding radius is approximately constant and carries limited semantic value, making the space amenable to magnitude normalization without significant semantic degradation.

\textbf{Magnitude Invariance and Directional Semantic Encoding.}
We validate the directional nature of the embedding space of all the diffusion-based text encoders through set of four experiments. First, we conduct a \textbf{magnitude scaling experiment} (Figure~\ref{fig:mag_and_nn}, right). Given an embedding $h \in \mathbb{R}^D$, we write it as $h = \|h\|_2 \cdot \tilde{h}$, where $\tilde{h}$ is the unit-normalized direction. By fixing the direction $\tilde{h}$ and scaling $\|h\|_2$ from $50\%$ to $200\%$ of its original norm, we observe that the generated images contain the same semantic content across all models (More results in Appendix. ~\ref{geo:textspace}.2). This confirms that $\|h\|_2$ is not a semantic carrier. Second, we perform a \textbf{nearest-neighbor (NN)} analysis in token space. As shown in Figure \ref{fig:mag_and_nn} (left), linear distance traversal retrieves semantically unrelated artifacts, such as single characters or random texts. Conversely, cosine similarity (angular NN) consistently retrieves coherent semantic neighbors. This provides direct evidence that semantic structure resides in direction, not magnitude in the text encoders. (More results in Appendix. ~\ref{geo:textspace}.3). Third, we evaluate \textbf{concept identification} across 150 visual
concepts and four diffusion backbones using embedding direction,
magnitude, and Euclidean distance. Concept embeddings are constructed
from ImageNet templates (\cite{openai_clip_imagenet_prompts}) and evaluated
on held-out prompts and LAION/COCO captions (\cite{schuhmann2021laion,lin2014microsoft}). Direction-based identification achieves 98.4-99.6\% accuracy, compared with 1.1-3.0\% using magnitude alone.
Fourth, we analyze \textbf{radial versus directional perturbations}
in image generation using DINOv2 similarity (\cite{oquab2023dinov2}),
source retention, and target acquisition. Radial changes better
preserve image appearance, while directional traversal achieves
88-100\% target acquisition.. Detailed quantitative results and the corresponding table for the third and fourth experiments are provided in Appendix. \ref{app:rebuttal_geometry}.

Together, these observations suggest that text encoder representations in diffusion models, including T5-based embeddings, are predominantly directional, with semantic relationships governed largely by angular structure rather than magnitude. This motivates normalizing embeddings onto the hypersphere $\mathbb{S}^{D-1}$ and performing geodesic traversal(\cite{turaga2010statistical}), which preserves the intrinsic geometry of the space and naturally enables controlled subject and attribute manipulation.

\subsection{Concept Representations as Kent Distributions}
\label{sec:concept_geometry}
Existing text-based control and erasure methods(\cite{biswas2025cure, biswas2025now, baumann2025continuous}) typically represent concepts as single directions or low-dimensional subspaces in embedding space. However, under the hyperspherical geometry established in Section~\ref{sec:hypersphere}, such representations fail to capture the full semantic variability of a concept. A concept (e.g., ``cat'') does not correspond to a single direction, but to a collection of embeddings induced by diverse semantic contexts (e.g., ``small cat'', ``sleeping cat'', ``cartoon cat''). We therefore model concepts as distributions on the hypersphere rather than isolated vectors, and investigate the geometric structure of these distributions.

To identify the appropriate distributional model, we collect concept embeddings using prompt templates (\cite{clip_prompts, radford2021learning} and evaluate three candidates: 
(i) the von Mises–Fisher (vMF) distribution (\cite{banerjee2005clustering}), which models an \emph{isotropic} spherical patch centered at a mean direction, 
(ii) mixtures of vMF (moVMF), which approximate multi-modal structure via multiple isotropic components, and 
(iii) the Kent distribution (\cite{kasarapu2015modelling}), which models an \emph{anisotropic} elliptical patch on the hypersphere. Empirically, we find that concept embeddings exhibit clear anisotropic structure: variance is not uniform in all directions, but aligned along dominant semantic axes (e.g., size, style, pose). As shown in Appendix. ~\ref{geo:textspace:concept}, the Kent distribution consistently provides the best fit across all concepts and encoders, exhibiting non-zero anisotropy. The Kent distribution is defined as:
\begin{equation}
p(x) \propto \exp\left( \kappa \mu^\top x + \beta \big[(\gamma_1^\top x)^2 - (\gamma_2^\top x)^2 \big] \right),
\end{equation}
where $\mu$ is the mean direction, $\kappa$ controls concentration, $\beta$ captures anisotropy, and $\gamma_1, \gamma_2$ define principal axes. Thus, geometrically, each concept corresponds to an elliptical patch on the hypersphere rather than a single point or direction. This perspective directly informs our control framework: instead of manipulating isolated semantic directions, we operate over Kent concept distributions, enabling more precise and structured subject and attribute control.

\subsection{Contamination in Token Representations}
\label{sec:token_contamination}
We now analyze how semantic information is distributed across token embeddings in diffusion model text encoders. Diffusion models operate on sequences of token embeddings(SD 1.5 Clip, SDXL Clip, SD3.5-medium T5, Flux-Schnell T5), where inter-token interactions play a central role in generation.
In CLIP-based encoders (SD1.5, SDXL), attention is causal, such that each token aggregates information only from preceding tokens. Consequently, the representation at position $p$ depends on the sequence prefix:
$h^{(p)} = f(x^{(1)}, x^{(2)}, \dots, x^{(p)})$
implying that downstream tokens ($p > p^*$) incorporate information from a preceding concept token at position $p\*$.
For example, in the prompt \textit{``a photo of a cat sitting on a chair''},
the representations of downstream tokens such as \textit{``sitting''},
\textit{``on''}, and \textit{``chair''} depend on the embedding of
\textit{``cat''}. As a result, modifying the concept token propagates changes beyond its local position into downstream token representations.

\begin{wrapfigure}{r}{0.6\textwidth}
\vspace{-10pt}
\centering
\includegraphics[width=\linewidth]{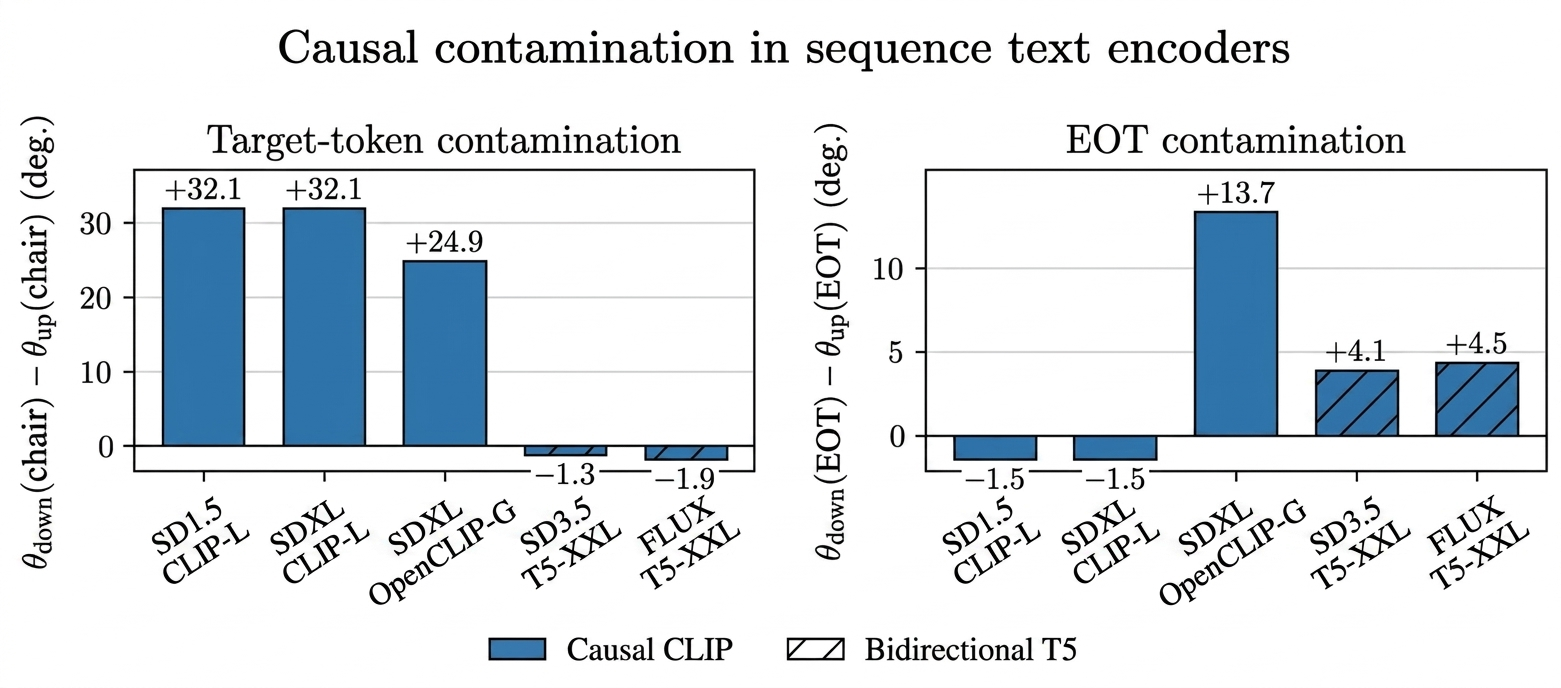}
\vspace{-10pt}
\caption{
Causal contamination across text encoders. We measure angular changes between
upstream and downstream prompt structures at the target token
(\emph{chair}, left) and the EOT token (right).
CLIP-based encoders show strong rightward semantic propagation, whereas
T5-based encoders exhibit more distributed bidirectional behavior.
}
\label{fig:causal_contamination}
\vspace{-8pt}
\end{wrapfigure}

To quantify this effect, we measure the angular separation between concept
embeddings (``cat'' vs.\ ``dog'') across token positions:
$\theta^{(p)} = \arccos \left( \tilde{h}_{c_1}^{(p)} \cdot \tilde{h}_{c_2}^{(p)} \right)$.
We compare two prompt structures: a \emph{downstream} setting
(``a photo of a [cat$\rightarrow$dog] sitting on a chair'') and an \emph{upstream} setting
(``a chair with a [cat$\rightarrow$dog] on it''), where the position of the concept token is reversed. As shown in Figure~\ref{fig:causal_contamination}, CLIP-based encoders with
sequence outputs exhibit strong positional asymmetry: Downstream tokens (``chair'') show significantly larger concept-dependent angular changes,
while upstream tokens remain largely invariant to the subject change. This indicates that concept identity propagates rightward through the sequence.

In contrast, T5-based encoders (SD3.5, FLUX), which use bidirectional attention, show weaker positional asymmetry (Fig.~\ref{fig:causal_contamination}). This means that modifying the concept token alters both the upstream and downstream tokens, hence the angular difference between the upstream token ``chair''  (Prompt: ``a photo of a cat sitting on a chair'') and downstream token ``chair'' (Prompt:``a chair with a cat on it'') is lower, than that of CLIP.
We also observe that the end-of-text (EOT) token shows non-trivial angular variation, reflecting its role as a global aggregation of sequence information. Changes in the concept token therefore influence not only local downstream tokens but also globally aggregated representations.

Contamination in token representations has important implications for text-based editing. Effective subject and attribute control must account for both token position and the underlying direction of information flow, which varies across diffusion model architectures. This motivates our approach to localize transformations to semantically meaningful token positions while respecting the directional geometry of the embedding space.

\section{HEART: Hyperspherical Embedding Alignment via Kent Traversal}
Building on the geometric and token-level analysis in Section \ref{sec:geo}, we propose
\textbf{HEART}, a training-free framework for controllable generation that
operates directly in text embedding space.
Our method performs \emph{structured, position-aware transformations} that selectively modify concept-relevant components while preserving global context. This is achieved through three steps: (1) estimating concept anchors from prompt templates, (2) isolating concept-specific directions from context-bearing tokens, and (3) performing controlled traversal in embedding space while accounting for token-level contamination. We form this framework for two tasks: \textbf{HEART-S} for subject control and \textbf{HEART-A} for attribute control.

\subsection{HEART-S: Subject Control}
Our goal is to perform subject replacement while preserving the
background and scene layout, without any retraining or modification
of diffusion model weights. Given an input prompt
$P = (\text{BOS}, x^{(1)}, \dots, x^{(T)}, \text{EOT}, \text{PAD}, \dots)$,
e.g., \textit{``a cat on a chair''}, say we aim to replace the subject
\textit{``\textcolor{red}{cat}''} with a target concept (say \textit{``\textcolor{cyan}{dog}''}).

\textbf{Step 1: Estimate concept anchors.}
We construct a small prompt pool for each concept and model it as a
Kent distribution on the hypersphere. The mean direction of this distribution provides the source and target anchors, $\mu_s$ and $\mu_t$, respectively.
\begin{figure*}[t]
\centering
\includegraphics[
    width=0.78\linewidth
]{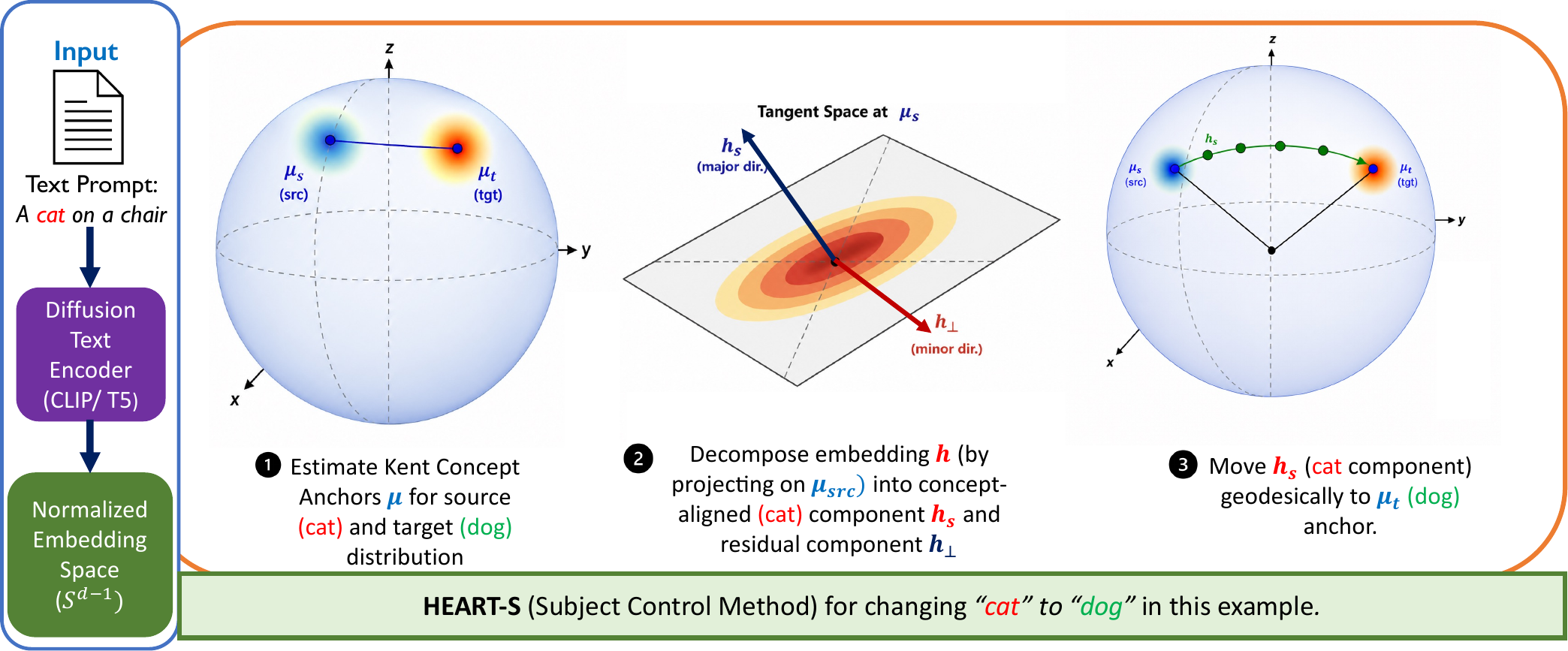}
\caption{
Overview of \textbf{HEART-S}. Visualization of geodesic traversal applied to the subject token and EOT token. The complete HEART-S pipeline (Section 3.1) additionally includes downstream token
disentanglement and contamination-aware semantic propagation.}
\label{fig:heart_method}
\end{figure*}

\textbf{Step 2: Edit the subject token.}
We first identify the subject token at position $p^*$ (e.g., ``\textcolor{red}{cat}'') and
normalize its embedding as $\tilde{h}^{(p^*)}=h^{(p^*)}/\|h^{(p^*)}\|_2$.
We then decompose it into a component aligned with the source anchor $\mu_s$
and a residual component:
\begin{equation}
\tilde{h}^{(p^*)}
=
\underbrace{\alpha_{p^*}\mu_s}_{h_{s}^{(p^*)}}
+
\underbrace{h_{\perp}^{(p^*)}}_{\text{context}},
\qquad
\alpha_{p^*}
=
\left\langle \tilde{h}^{(p^*)}, \mu_s \right\rangle .
\end{equation}
Here, $h_{s}^{(p^*)}$ captures the source-concept-aligned component, while $h_{\perp}^{(p^*)}$ preserves the residual context. We replace only the source-aligned direction by geodesically traversing from the source anchor $\mu_s$ toward the target anchor $\mu_t$ on the hypersphere:
$\mu_{s\rightarrow t}
=
\mathrm{Geo}(\mu_s,\mu_t;\lambda)$,
where $\lambda$ controls edit strength. The edited subject embedding is then
reconstructed as
$\hat{h}^{(p^*)}
=
\mathrm{Norm}\left(
\alpha_{p^*}\mu_{s\rightarrow t}
+
h_{\perp}^{(p^*)}
\right)$,
and rescaled to the original embedding norm.

\textbf{Step 3: Account for global tokens (EOT and PAD).}
Due to token contamination (Section~2.3), concept information is also
present in the end-of-text (EOT) and padding (PAD) tokens, which act as
global scene summaries. Therefore, modifying only the subject token is
insufficient. We construct corresponding anchors for these global tokens
and apply the same directional replacement to their concept-aligned
components.

\textbf{Step 4: Edit contaminated context tokens.}
Even after modifying the subject and global tokens, residual concept information persists in surrounding tokens (\textit{"on", "a", "chair"} in our example prompt in Fig ~\ref{fig:heart_method}) due to contamination.
We therefore extend the transformation to downstream tokens. Since the degree of contamination varies across positions, we apply the
edit \emph{proportionally}, scaling the transformation strength based on
each token’s distance from the subject token in embedding space.

\textbf{Step 5: Delayed injection for stable background formation.}
For CLIP-based encoders, upstream tokens do not show strong causal contamination in the text encoder, but they still influence generation through cross-attention in the diffusion model. To prevent disruption of background formation, we inject the modified embeddings only after an initial fraction (typically $10$-$15\%$) of the denoising steps. This allows the background to stabilize before applying subject edits.
\begin{figure*}[tbp]
\centering
\vspace{-8pt}
\includegraphics[width=0.8\linewidth]{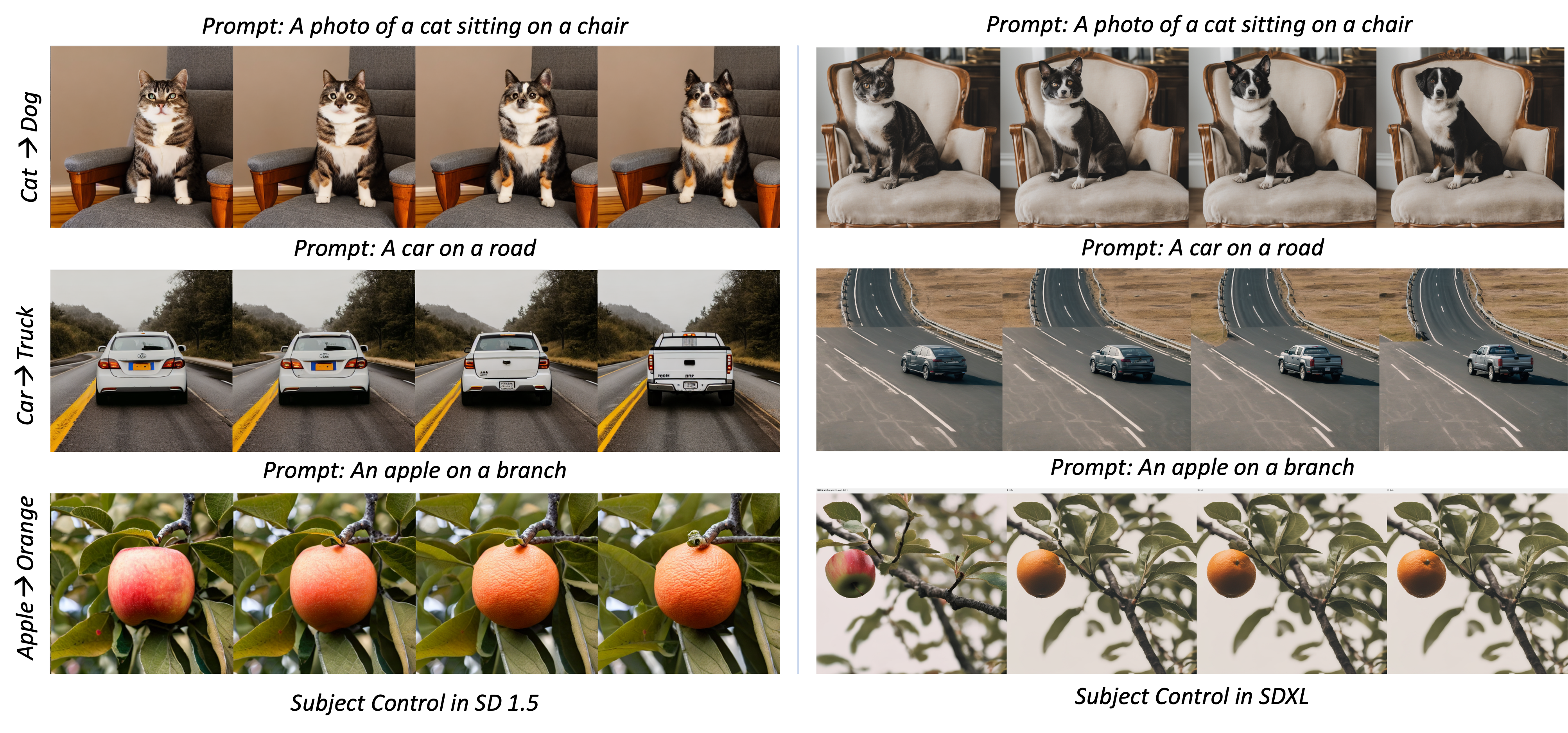}
\caption{
\textbf{HEART-S enables consistent subject replacement with minimal background drift.}
We show results for SD1.5 (left) and SDXL (right) across three subjects:
\textit{cat$\rightarrow$dog}, \textit{car$\rightarrow$truck}, and \textit{apple$\rightarrow$orange}.
The transformation is controlled by the angular parameter $\lambda$.
}
\vspace{-10pt}
\label{fig:subject_control_main1}
\end{figure*}

We demonstrate subject control results in Fig.~\ref{fig:subject_control_main1} on SD1.5 and SDXL. Additional results across other diffusion models are provided in the Appendix.

\subsection{HEART-A: Attribute Control}
\begin{figure*}[t]
    \centering
    \includegraphics[width=0.78\textwidth]{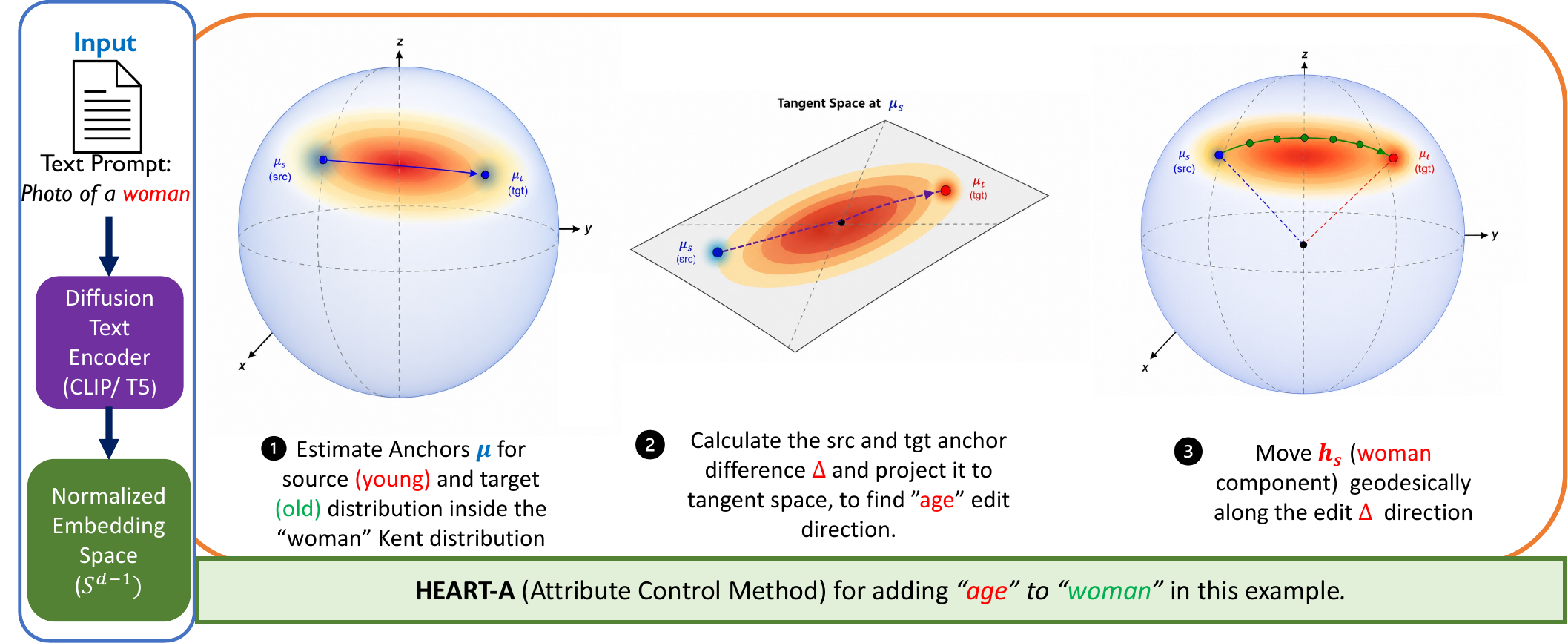}
    \vspace{-2mm}
    \caption{
Overview of \textbf{HEART-A}. Attribute control is performed by estimating
attribute anchors on the normalized embedding hypersphere, projecting the
edit direction into the local tangent space, and performing geodesic
traversal within the corresponding Kent distribution. Additional EOT-token
and downstream-token edits are described in Section~3.2.
    }
    \label{fig:heart_attribute_control}
    \vspace{-3mm}
\end{figure*}

\begin{figure*}[t]
    \centering
    \includegraphics[width=0.8\textwidth]{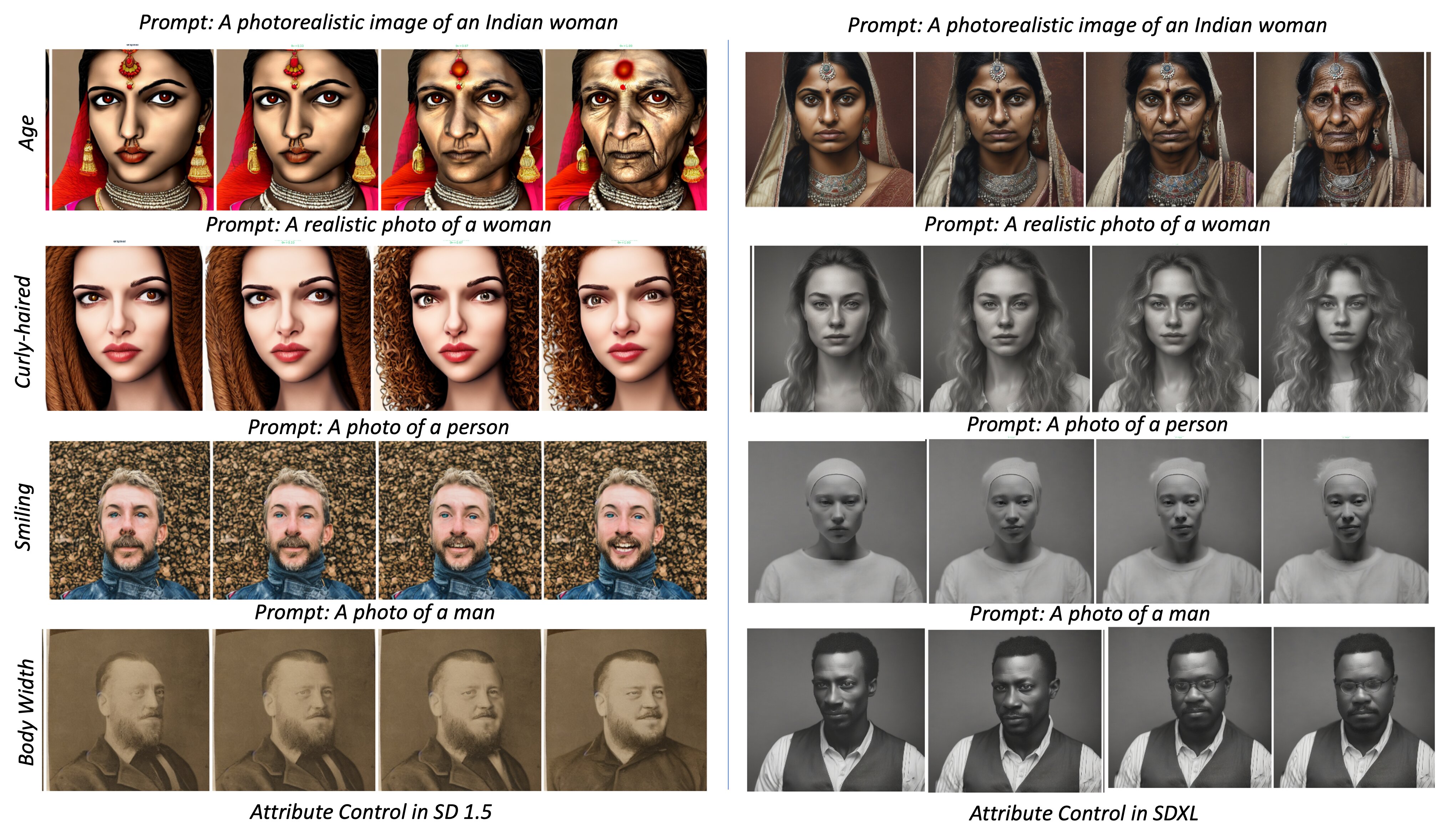}
    \vspace{-3mm}
    \caption{
    \textbf{HEART-A attribute control across SD\,1.5 and SDXL.}
    Left: SD\,1.5 results.
    Right: SDXL results.
    HEART-A enables smooth geodesic traversal for diverse semantic attributes,
    including age, hair texture, facial expression, and body shape.
    Across both diffusion backbones, the edits remain monotonic while largely
    preserving subject identity, pose, and scene structure.
    }
    \label{fig:attribute_control_main}
    \vspace{-2mm}
\end{figure*}
While HEART-S performs subject replacement, attribute control requires
localized manipulation within a concept manifold without altering subject
identity or scene structure. From Section~\ref{sec:concept_geometry}, we know that each concept is represented not as a single direction, but as a Kent distribution (elliptical patch) on the hypersphere. Variations such as age, smile, hair texture, or body shape are encoded as structured directional changes within this local concept geometry. HEART-A therefore models attribute control as a controlled geodesic traversal within a concept’s Kent patch, enabling fine-grained semantic edits while preserving the surrounding concept structure.

\textbf{Step 1: Attribute-conditioned concept geometry.}
Given a concept $c$ and an attribute pair
$(a^{-}, a^{+})$ (e.g., \textit{\textcolor{red}{young}} $\leftrightarrow$ \textit{\textcolor{cyan}{old}}), we construct two prompt pools:
a base pool conditioned on $a^{-}$ and a target pool conditioned on $a^{+}$.
For each prompt, we extract the normalized embedding at the subject token: $\tilde{h}^{(i)} \in \mathbb{S}^{D-1}$. Using the procedure described in Section~\ref{sec:concept_geometry}, we estimate Kent distributions for both pools and obtain their mean directions: $\mu_{a^-}, \mu_{a^+}$. These anchors define the local semantic geometry of the attribute transition
around the concept distribution.

\textbf{Step 2: Tangent-space attribute direction.}
A naive Euclidean difference between attribute anchors introduces radial
components that violate the hyperspherical structure of the embedding space.
We therefore compute the anchor displacement
$\Delta = \mu_{a^+} - \mu_{a^-}$ and project it onto the tangent space at
$\mu_{a^-}$:
$
\Delta^{\mathrm{tan}}
=
\Delta
-
(\Delta^\top \mu_{a^-})\mu_{a^-}.
$
The normalized tangent vector
$
d_a
=
\frac{\Delta^{\mathrm{tan}}}{\|\Delta^{\mathrm{tan}}\|_2}
$
defines the attribute direction inside the local Kent patch.

\textbf{Step 3: Geodesic attribute traversal.}
Given a token embedding $h^{(p)}$, we first normalize it as
$\tilde{h}^{(p)} = h^{(p)}/\|h^{(p)}\|_2$.
We then move $\tilde{h}^{(p)}$ along the attribute direction $d_a$ on the
hypersphere:
$
\hat{h}^{(p)}
=
\mathrm{Exp}_{\tilde{h}^{(p)}}\!\left(\lambda d_a\right),
$
where $\mathrm{Exp}$ denotes the exponential map on $\mathbb{S}^{D-1}$ and
$\lambda$ controls edit intensity. Equivalently, this corresponds to a
geodesic step from the current token embedding along the tangent attribute
direction as shown in Figure ~\ref{fig:heart_attribute_control}. The edited embedding is then rescaled to its original norm before
being passed to the diffusion model.

\textbf{Step 4: Contamination-aware attribute propagation.}
As shown in Section~\ref{sec:token_contamination},
attribute information propagates beyond the concept token into downstream and
global tokens through attention.
Editing only the nominal attribute token therefore produces incomplete or
unstable transformations. To address this, HEART-A applies the edit not only to the subject token, but also to downstream and EOT tokens.
The edit strength is scaled according to the token’s semantic proximity to
the concept token: $\lambda_p = \lambda \cdot w(p)$, where $w(p)$ decays with token distance in embedding space.
This produces localized attribute modifications while preserving scene layout, identity, and non-edited context. Representative qualitative results are shown in Figure~\ref{fig:attribute_control_main}. Additional results across other diffusion models are provided in the Appendix.
\vspace{-10pt}
\section{Experimental Results}
\vspace{-5pt}
We evaluate HEART across three semantic control settings:
(1) subject control, (2) attribute control, and
(3) challenging editing scenarios involving large semantic shifts,
object removal, artistic style transfer, and compositional prompts.
Experiments are conducted primarily on SDv1.5 and SDXL, with
additional validation on SD3.5-medium and FLUX.
For subject control, we evaluate 80 source-target concept pairs
spanning animals, natural entities, objects, vehicles, food,
clothing, and furniture, using three prompts and three random
seeds per pair. For attribute control,  we evaluate 20 attributes
covering both human and non-human concepts, including changes
in appearance, age, expression, and other semantic properties.
We compare HEART against diffusion-space editing methods and
recent text-based semantic control approaches. Additional implementation details, prompt templates, ablations, cross-model comparisons, and extended qualitative results are provided in the Appendix. 

\noindent\textbf{Evaluation Metrics.}
Conventional image-editing evaluations rely on CLIP-based
text-image similarity (\cite{hessel2021clipscore}) to assess semantic
alignment and full-image LPIPS (\cite{zhang2018unreasonable}) to measure perceptual changes. However, CLIP similarity does not capture whether the surrounding scene is preserved, while full-image LPIPS  conflates intended
subject modifications with unintended background changes.
Thus, these metrics alone \underline{cannot} adequately assess the trade-off between editing effectiveness and scene preservation.

To address this, we adopt \textbf{\emph{region-aware evaluation}} using
Grounded-SAM (\cite{ren2024grounded}) to separate foreground and background regions. Foreground LPIPS (FG-LPIPS) measures the perceptual change
within the edited subject region, where higher values indicate
stronger modification, whereas background LPIPS (BG-LPIPS)
measures unintended changes outside the subject region,
where lower values indicate better preservation.
We further assess background structural and semantic
consistency using masked SSIM (\cite{wang2004image})  and background DINOv2
similarity (BG-DINO) (\cite{oquab2023dinov2}), respectively, where higher values indicate better preservation. Editing success is evaluated using target-concept accuracy, DINOv2 classification accuracy,
and CLIP-T.

\subsection{Subject Control Evaluation}

\begin{table*}[t]
\centering
\scriptsize
\setlength{\tabcolsep}{3pt}
\renewcommand{\arraystretch}{1.15}

\resizebox{\textwidth}{!}{%
\begin{tabular}{lllcccccccc}
\toprule
\textbf{Method / Runtime (s)}
& \textbf{Space}
& \textbf{Inv.}
& \shortstack{\textbf{Target}\\\textbf{Acc.} $\uparrow$}
& \shortstack{\textbf{DINO}\\\textbf{Acc.} $\uparrow$}
& \shortstack{\textbf{Source}\\\textbf{Persist.} $\downarrow$}
& \textbf{CLIP-T} $\uparrow$
& \textbf{FG-LPIPS} $\uparrow$
& \textbf{BG-LPIPS} $\downarrow$
& \shortstack{\textbf{Masked}\\\textbf{SSIM} $\uparrow$}
& \textbf{BG-DINO} $\uparrow$ \\
\midrule

\rowcolor[gray]{0.92}
Direct Swap / \textbf{2.95}
& Prompt & No
& \textbf{0.977} & \textbf{0.988} & \textbf{0.018}
& \textbf{0.294} & \textbf{0.683} & 0.504
& 0.536 & 0.645 \\

\rowcolor[gray]{0.92}
LEDITS++ [\cite{brack2024ledits++}] / 31.68
& Attn. & Yes
& 0.836 & 0.813 & 0.105
& 0.285 & 0.540 & 0.221
& 0.795 & 0.781 \\

SEGA [\cite{brack2023sega}] / 4.65
& $\epsilon_\theta$ & No
& 0.719 & 0.696 & 0.269
& 0.277 & 0.430 & \underline{0.180}
& \underline{0.846} & \underline{0.843} \\

\midrule

\textbf{HEART-S} / \underline{2.99}
& $\mathbb{S}^{D-1}$ & No
& 0.621 & 0.614 & 0.544
& 0.255 & 0.421 & \textbf{0.137}
& \textbf{0.884} & \textbf{0.876} \\

\textbf{HEART-S (earlier injection)} / \underline{2.99}
& $\mathbb{S}^{D-1}$ & No
& \underline{0.936} & \underline{0.953} & \underline{0.023}
& \underline{0.291} & \underline{0.680} & 0.504
& 0.536 & 0.659 \\

\bottomrule
\end{tabular}%
}

\caption{
\textbf{Subject control on SDXL.}
HEART-S achieves the \textbf{best background preservation} (BG-LPIPS,
masked SSIM, BG-DINO) with competitive subject modification
(FG-LPIPS). Earlier injection improves target acquisition to
near Direct Swap performance, at the cost of background preservation.
Note, \textit{Direct Swap simply regenerates images using the target prompt
and is not an editing method.} HEART achieves the \textbf{lowest runtime}
among editing baselines without inversion or optimization.
Further details are discussed in Section~4.1.
\textbf{Bold} and \underline{underlined} denote the best and
second-best results, respectively.
}
\vspace{-10pt}
\label{tab:subject_control}
\end{table*}

We first evaluate training-free subject replacement, where the goal is to
modify a source concept $c_{\mathrm{src}}$ into a target concept
$c_{\mathrm{tgt}}$ while preserving scene structure and contextual content.
We compare HEART against LEDITS++~(\cite{brack2024ledits++}),
which relies on inversion and attention-based editing, and
SEGA~(\cite{brack2023sega}), which operates through semantic guidance
in the diffusion noise-prediction space ($\epsilon_\theta$).
As shown in Table~\ref{tab:subject_control}, HEART achieves the
best background preservation among the evaluated methods,
while allowing localized subject modifications.
Although LEDITS++ and SEGA attain higher target accuracy,
they introduce greater background changes.
The trade-off between editing strength and scene consistency can be controlled through the denoising intervention
step: applying HEART earlier increases target accuracy from
0.621 to 0.936, approaching Direct Swap, but at the
cost of reduced background preservation.
Note, Direct Swap is a regeneration baseline that
simply replaces the source concept in the prompt rather than
editing an existing generation. Furthermore, HEART achieves the lowest runtime among editing
methods (2.99\,s), compared with SEGA (4.65\,s) and
LEDITS++ (31.68\,s), without requiring inversion or optimization.

\subsection{Attribute Control Evaluation}












\begin{table*}[t]
\centering
\scriptsize
\setlength{\tabcolsep}{3pt}
\renewcommand{\arraystretch}{1.12}

\resizebox{\textwidth}{!}{%
\begin{tabular}{llccccccc}
\toprule
\textbf{Method}
& \textbf{Training}
& \shortstack{\textbf{Mem.}\\\textbf{(GB)}}
& $\boldsymbol{\Delta}$\textbf{CLIP} $\uparrow$
& \shortstack{\textbf{Full}\\\textbf{LPIPS} $\downarrow$}
& \textbf{FG-LPIPS} $\uparrow$
& \textbf{BG-LPIPS} $\downarrow$
& \textbf{FG/BG} $\uparrow$
& \shortstack{\textbf{Avg. Total Cost}\\\textbf{/ Attribute} $\downarrow$} \\
\midrule

\rowcolor[gray]{0.92}
Prompt Rewrite
& No & N/A
& \textbf{+0.062} & 0.477
& \underline{0.365} & 0.341
& 1.1 & N/A \\

Concept Sliders\cite{gandikota2024concept}
& Yes & 3.87
& \underline{+0.049} & 0.531
& \textbf{0.402} & 0.396
& 1.0 & 0.965 h (57.9 min) \\

Text Sliders\cite{chiu2026text}
& Yes & 5.68
& +0.018 & \underline{0.087}
& 0.089 & \underline{0.070}
& 1.3 & \underline{0.209 h (12.5 min)} \\

Attribute Control\cite{baumann2025continuous}
& Yes & 24.66
& +0.026 & 0.113
& 0.143 & 0.067
& \textbf{2.1} & 7.31 h (439 min) \\

\midrule

\textbf{HEART-A}
& \textbf{None} & $\sim$0
& +0.011 & \textbf{0.061}
& 0.077 & \textbf{0.045}
& \underline{1.7}
& \textbf{0.083 h (5.0 min)} \\

\bottomrule
\end{tabular}%
}

\caption{
\textbf{Attribute control on SDXL.}
HEART-A achieves the \textbf{lowest} full-image and background LPIPS,
indicating strong scene preservation, while enabling localized
attribute changes without training. Although optimization-based
methods achieve larger semantic changes, HEART-A operates at the \textbf{lowest} computational cost. Prompt Rewrite directly
regenerates images using modified prompts and serves only as
a reference baseline. \textbf{Bold} and \underline{underlined} values indicate the best and second-best results, respectively. More explained in Section 4.2.
}
\vspace{-10pt}
\label{tab:attribute_control}
\end{table*}

We next evaluate continuous attribute manipulation across 20 attributes
spanning human and non-human concepts, including age, facial expression,
hair texture, body shape, and object appearance. We compare HEART-A
against Concept Sliders (\cite{gandikota2024concept}),
Text Sliders (\cite{chiu2026text}), and Attribute Control (\cite{baumann2025continuous}). These methods rely on finetuning or optimization to learn attribute-specific editing directions, whereas HEART-A performs geodesic traversal directly in the text embedding space without
any model training. Conventional metrics such as $\Delta$CLIP and full-image LPIPS measure semantic and perceptual changes but cannot distinguish
intended edits from unintended scene modifications. We therefore use foreground LPIPS to assess attribute changes and background LPIPS to evaluate scene preservation. As shown in Table~\ref{tab:attribute_control}, HEART-A achieves the lowest BG-LPIPS (0.045) and full-image LPIPS (0.061), demonstrating superior preservation among the evaluated methods. While trained baselines achieve larger $\Delta$CLIP and foreground changes, these are often accompanied by substantial modifications to the surrounding scene, indicating a less favorable editing-preservation trade-off. In contrast, HEART-A enables
more localized attribute control with the best overall and
background preservation, while requiring only 5 minutes per attribute without finetuning or optimization.

\subsection{Challenging Editing Scenarios}
\begin{figure*}[t]
    \centering
    \includegraphics[width=0.8\textwidth]{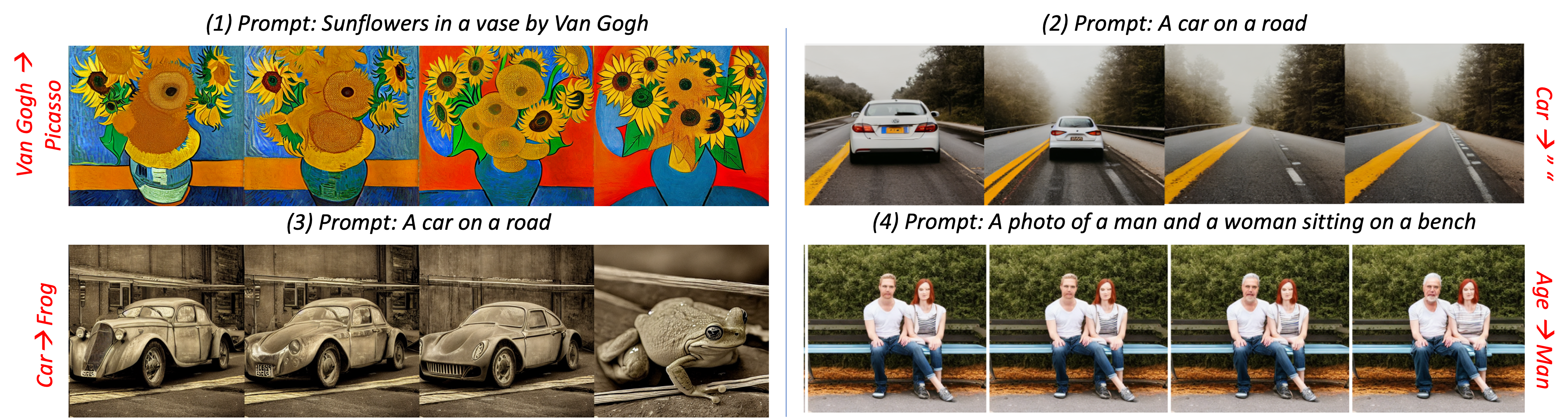}
    \vspace{-3mm}
    \caption{
    \textbf{Challenging editing scenarios with HEART.}
    Top-left: artistic style transfer from
    \textit{Van Gogh $\rightarrow$ Picasso}.
    Top-right: object removal from
    \textit{``a car on a road''}.
    Bottom-left: large semantic traversal from
    \textit{car $\rightarrow$ frog}.
    Bottom-right: localized compositional attribute editing where only the
    age of the man is modified while preserving the appearance of the woman
    and surrounding scene.
    }
    \label{fig:challenging_cases}
    \vspace{-2mm}
\end{figure*}

Beyond standard subject and attribute control, we evaluate HEART under several challenging editing settings that are difficult for existing
text-based methods in Figure ~\ref{fig:challenging_cases}.

\textbf{(1) Artistic style transfer.}
We first check artistic style manipulation using transitions such as
\textit{Van Gogh $\rightarrow$ Picasso}. Despite the substantial stylistic shift in color distribution, texture, and brushstroke structure, HEART preserves scene layout and object identity while producing smooth global style transitions.
\textbf{(2) Object removal.}
We further evaluate object erasure by traversing away from a target concept
direction while preserving contextual tokens.
For example, given prompts such as
\textit{``a car on a road''},
HEART suppresses the subject concept while retaining road structure,
background layout, and overall scene coherence.
\textbf{(3) Large semantic shifts.}
We consider extreme subject transitions such as
\textit{car $\rightarrow$ frog}, where the source and target concepts differ substantially in geometry, texture, and semantic category.
Despite the large angular separation between concepts, HEART maintains
background consistency while successfully transitioning the subject.
\textbf{(4) Compositional multi-subject control.}
We finally evaluate localized edits in prompts containing multiple subjects, such as \textit{``a photo of a man and a woman sitting on a bench''}. The goal is to modify only one subject (e.g., aging the man) while preserving the other subject and overall scene composition. HEART achieves localized control with minimal cross-subject interference.

\subsection{Additional Analysis and Ablations.}
We further validate the key design choices of HEART.
First, we examine the role of concept distribution geometry in HEART-S and HEART-A, comparing anisotropic Kent and isotropic vMF representations to assess their influence on subject and attribute control
(Appendix~\ref{app:rebuttal_kent}).
Second, we investigate token-level contamination and show
how correcting propagated concept information affects
editing accuracy and background preservation
(Appendix~\ref{app:rebuttal_tokens}).
Third, we analyze the denoising intervention step,
demonstrating its role in balancing semantic modification
and background consistency
(Appendix~\ref{app:rebuttal_injection}).
We also evaluate the robustness of concept anchor estimation
to prompt-template number and diversity
(Appendix~\ref{app:rebuttal_anchor}).
Finally, we provide cross-architecture evaluations of subject
and attribute control across diffusion backbones
(Appendices~\ref{app:rebuttal_subject}
and~\ref{app:rebuttal_attributes}), with detailed
experimental settings in Appendix~\ref{app:experimental_setup}.

\section{Conclusion}
We presented \textbf{HEART}, a geometry-driven framework for controllable T2I generation that revisits the structure of diffusion text embedding space. Across CLIP- and T5-based diffusion architectures, we showed that semantic representations are inherently directional and are better modeled as anisotropic Kent distributions on a hypersphere rather than linear Euclidean directions. We further showed that token-level contamination contributes to semantic entanglement during editing. Leveraging these geometric insights, HEART performs training-free geodesic traversal directly on the hyperspherical manifold, enabling precise subject replacement and fine-grained attribute control while preserving scene structure and identity. Unlike prior approaches, HEART requires no finetuning, inversion, or optimization, and generalizes across SD1.5, SDXL, SD3.5, and FLUX. Overall, our results suggest that controllable generation can emerge naturally from respecting the intrinsic geometry of representation space.
\newpage
\section*{Acknowledgment}
This work was supported in part by the Center for the Co-Design of Cognitive Systems (CoCoSYS), a research center under the Joint University Microelectronics Program (JUMP) 2.0, a Semiconductor Research Corporation (SRC) initiative sponsored by DARPA, and by National Science Foundation (NSF).

\bibliographystyle{plainnat}
\bibliography{neurips_2026}

\newpage


\appendix

\section{Geometry of Text Encoder Space in Diffusion Models}
\label{geo:textspace}
\subsection{Thin-shell structure of embedding space.}
To characterize the geometry of text encoder outputs, we examine whether embeddings lie on a thin hyperspherical shell. 
Intuitively, a thin shell implies that the norm of embeddings varies little across inputs, so representations are primarily distinguished by direction rather than magnitude.

\textbf{Metric.}
We quantify thin-shell behavior using the coefficient of variation of embedding norms:
\begin{equation}
\text{Thinness} = \frac{\mathrm{std}(\|h\|_2)}{\mathrm{mean}(\|h\|_2)},
\end{equation}
where $h \in \mathbb{R}^D$ denotes a token-level hidden state. 
Lower values indicate stronger concentration around a fixed radius (thin shell), while higher values indicate greater variability in magnitude.

\textbf{Experimental setup.}
We extract token-level hidden states from text encoders across four diffusion architectures:
SD1.5 (CLIP-L), SDXL (CLIP-L, OpenCLIP-G), SD3.5 (CLIP-L, CLIP-G, T5-XXL), and FLUX (CLIP-L, T5-XXL). 
Embeddings are computed over a diverse set of prompts (LAION-style and template-based), and statistics are aggregated across all valid tokens.

\textbf{Results.}
Table~\ref{tab:thinness_results} reports the thinness metric for all encoders.

\begin{table}[h]
\centering
\small
\begin{tabular}{l|l|c}
\toprule
Model & Encoder & Thinness \\
\midrule
SD1.5 & CLIP-L (seq) & 0.127 \\
SDXL & CLIP-L (seq) & 0.127 \\
SDXL & OpenCLIP-G (seq) & 0.080 \\
SD3.5 & CLIP-L (pool) & 0.017 \\
SD3.5 & CLIP-G (pool) & 0.057 \\
SD3.5 & T5-XXL (seq) & 0.227 \\
FLUX & CLIP-L (pool) & 0.017 \\
FLUX & T5-XXL (seq) & 0.221 \\
\bottomrule
\end{tabular}
\caption{
Thinness of embedding norms across text encoders. Lower values indicate stronger concentration on an ellipsoidal shell.
}
\label{tab:thinness_results}
\end{table}

CLIP-based encoders consistently exhibit low thinness ($\approx$ 0.02-0.13), indicating strong concentration on a hyperspherical shell. 
In contrast, T5-XXL exhibits significantly higher thinness ($\approx$ 0.22), reflecting weaker norm concentration. This distinction suggests that hyperspherical modeling is well-suited for CLIP-based encoders, where magnitude varies little and direction dominates.  For T5-based encoders, the embedding space remains structured but does not exhibit the same degree of spherical concentration.

\subsection{Magnitude Invariance under Radial Scaling: Qualitative}
\label{geo:textspace:mag}
\begin{figure*}[htbp]
\centering
\includegraphics[width=\linewidth]{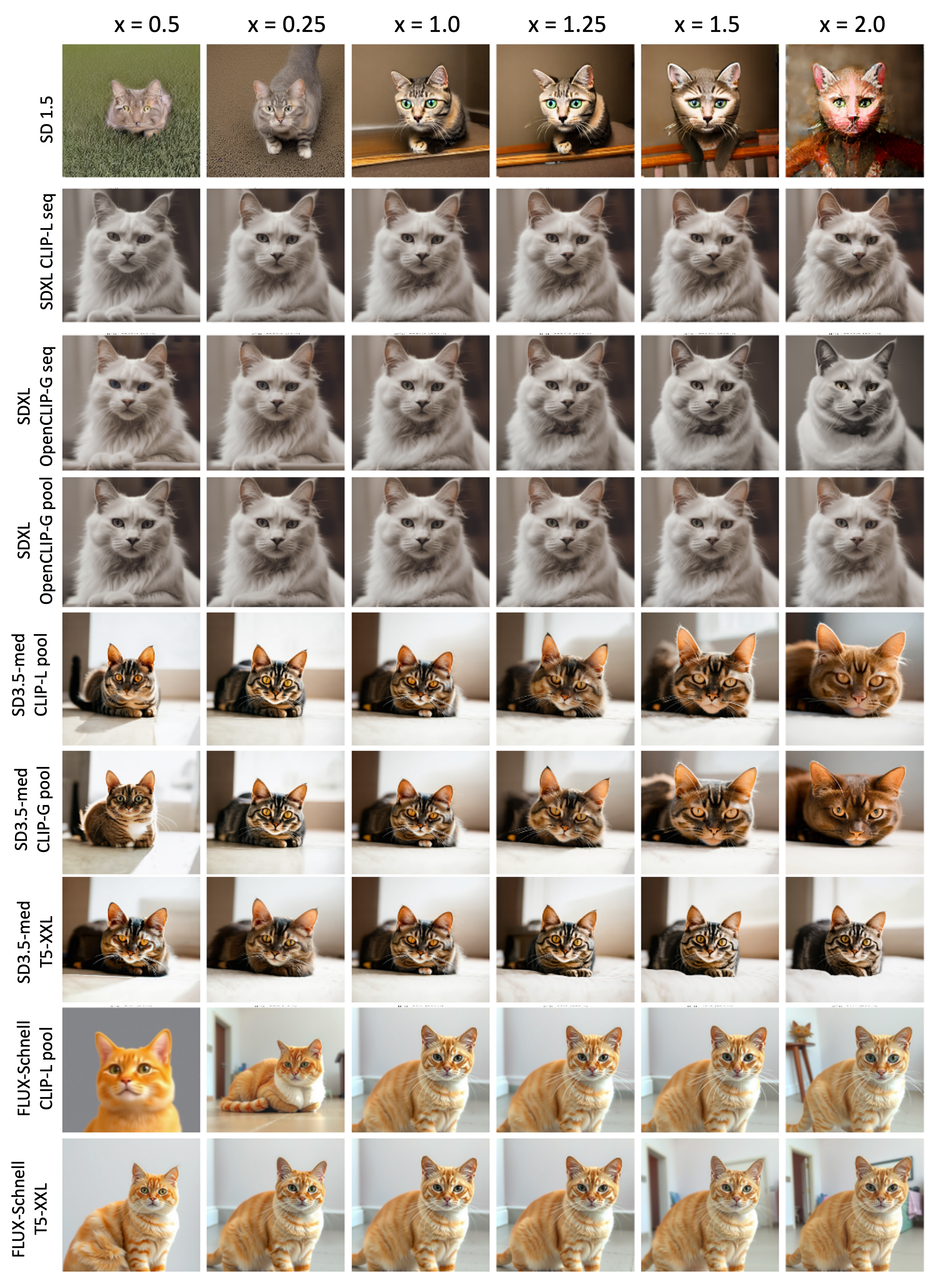}
\caption{
Magnitude scaling across text encoders. Each row corresponds to a model and prompt, while columns represent scaling factors $x \in [0.5, 2.0]$. 
Across CLIP and T5 encoders, semantic content remains consistent under large changes in embedding norm. 
}
\label{fig:mag_variation_appendix}
\end{figure*}
\textbf{Setup.}
To evaluate the role of embedding magnitude, we perform a controlled magnitude scaling experiment across all text encoders. 
Given a text embedding $h \in \mathbb{R}^D$, we decompose it as $h = \|h\|_2 \cdot \tilde{h}$ and construct modified embeddings of the form
\begin{equation}
h' = \alpha \|h\|_2 \cdot \tilde{h}, \quad \alpha \in \{0.5, 0.75, 1.0, 1.25, 1.5, 2.0\}.
\end{equation}
This procedure preserves direction while scaling magnitude. The modified embeddings are directly used as conditioning inputs to the diffusion model without any further adjustment.

\textbf{Models.}
We evaluate across all target architectures:
SD1.5 (CLIP-L), SDXL (CLIP-L, OpenCLIP-G), SD3.5 (CLIP-L/G, T5-XXL), and FLUX (CLIP-L, T5-XXL).

\textbf{Results.}
Figure~\ref{fig:mag_variation_appendix} shows representative generations across scaling factors. Across all CLIP-based and T5 encoders, the generated images remain visually consistent in terms of subject identity over a wide range of magnitude scaling. Even under large perturbations (e.g., $0.5\times$ to $2.0\times$), overall systematic semantic content remains stable.

\subsection{Nearest-Neighbor Analysis in Token Embedding Space}
\label{geo:textspace:nn}
To further validate that semantic information is encoded in the \emph{direction} rather than the \emph{magnitude} of text embeddings, we perform a nearest-neighbor (NN) retrieval analysis under two metrics: linear distance and angular similarity.

Given an embedding $h \in \mathbb{R}^D$, we decompose it as $h = \|h\|_2 \cdot \tilde{h}$, where $\tilde{h}$ is the unit-normalized direction. We then retrieve nearest neighbors from the token embedding vocabulary under:
\begin{itemize}
    \item \textbf{Linear distance:} $d_{\text{lin}}(h_i, h_j) = \|h_i - h_j\|_2$
    \item \textbf{Angular similarity:} $s_{\text{ang}}(h_i, h_j) = \frac{h_i^\top h_j}{\|h_i\|_2 \|h_j\|_2}$
\end{itemize}

Table~\ref{tab:nn_all_models} reports nearest neighbors for a diverse set of query tokens. We observe that linear distance often begets semantically incoherent tokens, including isolated characters or fragmented subwords. In contrast, angular similarity consistently provides semantically meaningful neighbors aligned with the query concept. This discrepancy arises because linear distance is not sensitive to the curved geometry of the token space, whereas angular similarity depends only on directional alignment. These results provide strong empirical evidence that semantic structure in text encoder space is primarily directional.

\begin{table*}[t]
\centering
\small
\setlength{\tabcolsep}{4pt}
\renewcommand{\arraystretch}{1.1}

\caption{
Nearest neighbors across text encoders. Angular similarity retrieves semantically coherent concepts, while linear distance retrieves magnitude-biased and often incoherent tokens.
}
\label{tab:nn_all_models}

\begin{tabular}{l|c|p{5.5cm}|p{5.5cm}}
\toprule
\textbf{Model (Encoder)} & \textbf{Query} & \textbf{Linear NN} & \textbf{Angular NN} \\
\midrule

\multirow{4}{*}{SD1.5 (CLIP-L)}
& dog 
& \texttt{A, B, C, X} 
& dog, dogs, puppy, pet \\

& truck 
& \texttt{T, R, K, \#} 
& truck, vehicle, van, transport \\

& apple 
& \texttt{A, B, C, 1} 
& apple, fruit, pear, banana \\

& chair 
& \texttt{C, H, R, !} 
& chair, seat, bench, furniture \\

\midrule

\multirow{4}{*}{SDXL (OpenCLIP-G)}
& dog 
& \texttt{teamsisd, bakhtawar, wakeupamerica} 
& dog, dogs, puppy, canine \\

& truck 
& \texttt{btsbbmas, swachhb, nswpol} 
& truck, pickup, vehicle, lorry \\

& apple 
& \texttt{randomhash, token123, junk} 
& apple, fruit, pear, peach \\

& chair 
& \texttt{zz, ctrl, junk} 
& chair, seat, sofa, furniture \\

\midrule

\multirow{4}{*}{FLUX (CLIP-L)}
& dog 
& \texttt{B, [id=197], ctrl-token} 
& dog, puppy, pet, animal \\

& truck 
& \texttt{A, J, B, misc} 
& truck, pickup, vehicle, van \\

& apple 
& \texttt{Z, Q, token, junk} 
& apple, fruit, pear, peach \\

& chair 
& \texttt{C, R, X, sym} 
& chair, seat, sofa, furniture \\

\midrule

\multirow{4}{*}{SD3.5 / FLUX (T5-XXL)}
& dog 
& \texttt{<extra\_id\_94>, [id=32122]} 
& dog, animal, pet, canine \\

& truck 
& \texttt{<extra\_id\_12>, [id=1123]} 
& truck, vehicle, transport, van \\

& apple 
& \texttt{<extra\_id\_33>, [id=778]} 
& apple, fruit, pear, peach \\

& chair 
& \texttt{<extra\_id\_7>, [id=445]} 
& chair, seat, furniture, stool \\

\bottomrule
\end{tabular}

\end{table*}

\subsection{Magnitude vs Radial Scaling: Quantitative}
\label{app:rebuttal_geometry}
\paragraph{Setup.}
We conduct two complementary experiments to quantitatively validate the
directional nature of text embeddings across 150 visual concepts.
First, for the embedding-level analysis, we construct a representative
embedding for each concept using ImageNet prompt templates
(\cite{radford2021learning, openai_clip_imagenet_prompts}). We then evaluate concept identification
using disjoint ImageNet templates and natural captions from LAION (\cite{schuhmann2021laion})  and
COCO (\cite{lin2014microsoft}), ensuring that the evaluation prompts are not used for constructing
the concept representations. We compare three representations:
(i) unit-normalized embedding directions, (ii) embedding magnitudes,
and (iii) raw Euclidean embeddings.

Second, for the generation-level analysis, we compare the effects of
radial and angular perturbations on generated images. Specifically,
we modify the magnitude of a text embedding while preserving its
direction, or rotate the embedding toward a target concept.
The perturbations are matched in Euclidean displacement to provide
a controlled comparison of their effects on image generation.

\paragraph{Models.}
We evaluate across four diffusion architectures: SD1.5 (CLIP-L),
SDXL (CLIP-L, OpenCLIP-G), SD3.5 (CLIP-L/G, T5-XXL),
and FLUX (CLIP-L, T5-XXL).

\paragraph{Metrics.}
For embedding-level evaluation, we report concept-identification
accuracy using direction and magnitude, together with the area under
the ROC curve (AUC) for cosine-based and radial discrimination.
For completeness, we also evaluate identification using raw Euclidean
embeddings. For generation-level evaluation, we measure the visual
similarity between original and modified generations using DINOv2
(\cite{oquab2023dinov2}), where higher similarity indicates greater
preservation of image appearance. We additionally report source-concept
retention which measures whether changing the embedding magnitude preserves the original semantic identity, and target-concept acquisition which measures whether rotating the embedding towards a target concept successfully changes the generated semantic quantity or not.

\begin{table*}[t]
\centering
\caption{Quantitative validation of directional semantics across diffusion
backbones.}
\label{tab:rebuttal_geometry_full}
\scriptsize
\setlength{\tabcolsep}{3pt}
\resizebox{\textwidth}{!}{%
\begin{tabular}{lccccccc}
\toprule
Model
& \shortstack{Concept identified\\from direction (\%) $\uparrow$}
& \shortstack{Concept identified\\from magnitude (\%) $\downarrow$}
& \shortstack{Cosine / Radial\\AUC $\uparrow$}
& \shortstack{Image similarity after\\magnitude change\\(DINOv2) $\uparrow$}
& \shortstack{Image similarity after\\target-direction change\\(DINOv2) $\downarrow$}
& \shortstack{Source retained (\%) \\ post magnitude change $\uparrow$}
& \shortstack{Target generated (\%) \\ post direction change $\uparrow$} \\
\midrule
SD1.5 & 99.1 & 3.0 & 0.989 / 0.644 & 0.436 & 0.028 & 60 & 88 \\
SDXL  & 99.6 & 1.1 & 0.993 / 0.612 & 0.746 & 0.014 & 99 & 100 \\
SD3   & 98.4 & 1.5 & 0.967 / 0.648 & 0.721 & 0.362 & 100 & 100 \\
FLUX  & 98.4 & 1.5 & 0.967 / 0.647 & 0.879 & 0.029 & 100 & 99 \\
\bottomrule
\end{tabular}%
}
\end{table*}

\paragraph{Results.}
Table~\ref{tab:rebuttal_geometry_full} shows that embedding direction
consistently captures semantic identity across all four diffusion
architectures, achieving 98.4-99.6\% concept-identification accuracy,
compared to only 1.1-3.0\% using embedding magnitude alone.
Similarly, cosine-based retrieval achieves substantially higher AUC
(0.967-0.993) than magnitude-based discrimination (0.612--0.648).
Identification using raw Euclidean embeddings also achieves
93.6-97.1\% accuracy, indicating that directional information
remains accessible without explicit normalization.

The generation-level analysis further supports these observations.
Magnitude perturbations generally preserve greater visual similarity
to the original generations, whereas directional perturbations
produce substantially larger visual changes. Importantly, rotating
embeddings toward target concepts achieves 88-100\% target-concept
acquisition across the four models. Together, these results indicate
that semantic identity is predominantly governed by embedding
direction, while changes in magnitude have a comparatively smaller
influence on the generated semantic content.

\subsection{Concept Distribution Modeling on the Hypersphere}
\label{geo:textspace:concept}

\textbf{Motivation.}
We model concepts as distributions on the hypersphere rather than single directions. 
To determine the appropriate representation, we compare three directional distributions: 
(i) von Mises-Fisher (vMF), (ii) mixtures of vMF (moVMF), and (iii) the Kent distribution.

\textbf{Distributions.}
The vMF distribution models an isotropic spherical patch centered at mean direction $\mu$ with concentration $\kappa$. 
moVMF extends this to multiple isotropic components. 
The Kent distribution generalizes vMF by modeling anisotropic (elliptical) structure:
\begin{equation}
p(x) \propto \exp\left( \kappa \mu^\top x + \beta \left[(\gamma_1^\top x)^2 - (\gamma_2^\top x)^2 \right] \right),
\end{equation}
where $\beta$ controls anisotropy and $\gamma_1, \gamma_2$ define principal axes.

\textbf{Model selection via BIC.}
We evaluate model fit using the Bayesian Information Criterion (BIC):
\begin{equation}
\text{BIC} = -2 \log \mathcal{L} + k \log N,
\end{equation}
where $\mathcal{L}$ is the likelihood, $k$ is the number of parameters, and $N$ is the number of samples. 
Lower BIC indicates a better trade-off between fit quality and model complexity.

\textbf{Results.}
Table~\ref{tab:kent_results} summarizes results across all concepts and encoders. 
Kent consistently achieves the lowest BIC across all concept-encoder pairs, with a $100\%$ selection rate, meaning in all cases Kent was found to be the best representative for concept distributions. The average anisotropy ratio is $\beta/\kappa \approx 0.12$, indicating consistent but moderate deviation from isotropy.

\begin{table*}[t]
\centering
\small
\caption{
Directional distribution fitting across text encoders. Kent consistently achieves the lowest BIC and captures anisotropic structure, while vMF and moVMF remain isotropic ($\beta=0$).
}
\label{tab:kent_results}

\begin{tabular}{l|c|c|c|c|c}
\toprule
\textbf{Encoder} 
& $\bar{r}$ 
& \textbf{BIC (vMF)} 
& \textbf{BIC (moVMF)} 
& \textbf{BIC (Kent)} 
& $\beta/\kappa$ (Kent) \\
\midrule

SD1.5 (CLIP-L)      
& 0.88-0.96 
& -1.2e4 
& -1.3e4 
& \textbf{-1.5e4} 
& 0.10-0.23 \\

SDXL (OpenCLIP-G)   
& 0.89-0.96 
& -2.0e4 
& -2.2e4 
& \textbf{-3.1e4} 
& 0.13-0.25 \\

SD3.5 (CLIP-G)      
& 0.80-0.88 
& -8.1e3 
& -8.6e3 
& \textbf{-9.9e3} 
& 0.10-0.14 \\

SD3.5 (CLIP-L)      
& 0.77-0.83 
& -4.2e3 
& -4.5e3 
& \textbf{-5.3e3} 
& 0.08-0.13 \\

FLUX (CLIP-L)       
& 0.78-0.83 
& -4.3e3 
& -4.6e3 
& \textbf{-5.4e3} 
& 0.08-0.13 \\

T5-XXL              
& 0.83-0.89 
& -2.8e4 
& -3.1e4 
& \textbf{-3.2e4} 
& 0.03-0.14 \\

\bottomrule
\end{tabular}
\end{table*}

These results show that concepts are best modeled as anisotropic directional distributions rather than single directions or isotropic clusters. 
This justifies our use of Kent-based representations for control operations.

\subsection{Role of Concept Distribution Geometry in HEART}
\label{app:rebuttal_kent}

\textbf{Motivation.}
In Section~\ref{sec:concept_geometry}, we establish that concept embeddings are better represented by anisotropic Kent distributions than isotropic vMF distributions. Here, we investigate how this distributional geometry influences the editing operations in HEART-S and HEART-A.

In HEART-S, subject replacement is primarily governed by the source and target concept anchors, $\mu_s$ and $\mu_t$. Since Kent and vMF estimate similar mean directions, replacing Kent with vMF has only a minor influence on subject control. In contrast, HEART-A performs attribute traversal within the local geometry of a concept distribution. The Kent distribution
captures direction-dependent variations through its anisotropy parameters, allowing the attribute traversal to account for the ellipsoidal neighborhood of the concept. An isotropic vMF distribution assumes equal spread in all tangent directions and therefore cannot capture these variations, potentially allowing edits to extend beyond the local concept neighborhood and introduce unintended changes.

\textbf{Experimental Setup.}
To isolate the contribution of anisotropic concept modeling,
we replace the Kent representation in HEART-A with an
isotropic vMF distribution while keeping all other editing
components unchanged. We evaluate five attributes spanning
human and non-human concepts: muscularity, smile, long hair,
surprised expression, and car age. We measure attribute
modification using $\Delta$CLIP and background preservation
using BG-LPIPS, where higher $\Delta$CLIP indicates stronger
semantic change and lower BG-LPIPS indicates better
background preservation.

\begin{table}[t]
\centering
\small
\setlength{\tabcolsep}{4pt}
\renewcommand{\arraystretch}{1.05}

\begin{minipage}{0.85\columnwidth}
\centering
\begin{tabular}{lcccc}
\toprule
\textbf{Attribute}
& \shortstack{\textbf{Kent}\\\textbf{BG-LPIPS} $\downarrow$}
& \shortstack{\textbf{vMF}\\\textbf{BG-LPIPS} $\downarrow$}
& \shortstack{\textbf{Kent}\\$\boldsymbol{\Delta}$\textbf{CLIP} $\uparrow$}
& \shortstack{\textbf{vMF}\\$\boldsymbol{\Delta}$\textbf{CLIP} $\uparrow$} \\
\midrule
Muscularity & \textbf{0.038} & 0.066 & 0.0055 & 0.0056 \\
Smile       & \textbf{0.044} & 0.075 & 0.0096 & 0.0161 \\
Long Hair   & \textbf{0.063} & 0.091 & 0.0049 & 0.0075 \\
Surprised   & \textbf{0.033} & 0.067 & 0.0020 & 0.0025 \\
Car Age     & \textbf{0.183} & 0.253 & 0.0397 & 0.0583 \\
\bottomrule
\end{tabular}

\vspace{3pt}
\caption{
\textbf{Kent versus vMF in HEART-A.}
Kent consistently improves background preservation
while retaining attribute control.
Bold denotes the better value.
}
\label{tab:kent_vmf_ablation}
\end{minipage}
\vspace{-20pt}
\end{table}

\textbf{Results.}
As shown in Table~\ref{tab:kent_vmf_ablation}, Kent achieves
lower BG-LPIPS than vMF across all five attributes,
indicating consistently better background preservation.
Although vMF yields larger $\Delta$CLIP values in some
cases, these stronger changes are accompanied by greater
background modifications. For example, in smile editing,
vMF increases $\Delta$CLIP from 0.0096 to 0.0161, but
also increases BG-LPIPS from 0.044 to 0.075.
Similarly, for car-age editing, vMF increases BG-LPIPS
from 0.183 to 0.253.

These results highlight the importance of concept
distribution geometry in HEART: while mean directions
are sufficient for estimating subject-replacement anchors
in HEART-S, the anisotropic Kent representation provides
a more suitable local geometry for attribute traversal
in HEART-A, improving the balance between attribute
modification and background preservation.

\subsection{Contamination-Aware Token Correction}
\label{app:rebuttal_tokens}

\textbf{Motivation.}
As discussed in Section~\ref{sec:token_contamination},
concept-specific information propagates beyond the subject
token to downstream and global token representations.
HEART accounts for this propagation by extending the
geometric transformation to the affected token positions.
Here, we investigate the contribution of these corrections
to subject replacement and background preservation.

\textbf{Experimental Setup.}
To isolate the effect of token-level correction, we ablate
the token positions modified by HEART while keeping the
diffusion backbone, prompts, random seeds, denoising
schedule, and editing strength fixed. We evaluate 47
source--target concept pairs with three seeds each
(141 matched generations per condition) on SD1.5,
which uses a single CLIP text encoder.
We compare subject-token-only editing with corrections
applied to end-of-text (EOT), padding (PAD), and
downstream tokens. Target accuracy measures successful
subject replacement, while foreground and background
LPIPS (FG-LPIPS and BG-LPIPS) quantify intended subject
changes and unintended scene modifications, respectively.
Subject masks are obtained using Grounded-SAM.

\begin{table}[t]
\centering
\scriptsize
\setlength{\tabcolsep}{4pt}
\renewcommand{\arraystretch}{1.05}
\begin{tabular}{llccc}
\toprule
\textbf{Cond.} & \textbf{Tokens Edited}
& \textbf{Acc.} $\uparrow$
& \textbf{BG-LPIPS} $\downarrow$
& \textbf{FG-LPIPS} $\uparrow$ \\
\midrule
C0 & None                       & 3\%  & --    & --    \\
C1 & Subject only               & 53\% & 0.099 & 0.309 \\
C2 & Downstream only            & 3\%  & 0.022 & 0.055 \\
C3 & Subject + EOT              & 54\% & 0.111 & 0.325 \\
C4 & Subject + EOT + PAD        & 65\% & 0.140 & 0.377 \\
C5 & Subject + EOT + PAD + Downstream
                                  & \textbf{72\%} & 0.157 & \textbf{0.406} \\
\bottomrule
\end{tabular}
\caption{
\textbf{Token correction ablation on SD1.5.}
Correcting propagated concept information progressively
improves subject replacement accuracy and foreground
modification, with an associated increase in background
changes. C5 denotes the complete HEART correction.
}
\label{tab:token_correction_ablation}
\end{table}

\textbf{Results.}
Table~\ref{tab:token_correction_ablation} shows that
editing only the subject token achieves 53\% target
accuracy. Incorporating EOT and PAD corrections
improves accuracy to 65\%, while the complete
contamination-aware correction achieves 72\%.
The corresponding FG-LPIPS increases from 0.309
to 0.406, indicating stronger intended subject
modification. In contrast, editing only downstream
tokens yields 3\% target accuracy, suggesting that
these representations contain propagated concept
information but are insufficient for subject
replacement independently.

The improved target acquisition is accompanied by
increased BG-LPIPS (0.099 to 0.157), highlighting
the trade-off between correcting propagated concept
information and preserving the surrounding scene.

\textbf{Rotation-Strength Ablation.}
To distinguish the contribution of token selection
from editing strength, we additionally vary the
rotation strength $\theta$ for subject-token-only
editing (C1) and the complete HEART correction (C5).

\begin{table}[t]
\centering
\scriptsize
\setlength{\tabcolsep}{5pt}
\renewcommand{\arraystretch}{1.05}
\begin{tabular}{ccccc}
\toprule
& \multicolumn{2}{c}{C1: Subject Only}
& \multicolumn{2}{c}{C5: HEART} \\
\cmidrule(lr){2-3}
\cmidrule(lr){4-5}
$\boldsymbol{\theta}$
& Acc. $\uparrow$ & BG-LPIPS $\downarrow$
& Acc. $\uparrow$ & BG-LPIPS $\downarrow$ \\
\midrule
0.50 & 15\% & 0.062 & 22\% & 0.092 \\
0.75 & 45\% & 0.090 & 60\% & 0.128 \\
1.00 & 53\% & 0.099 & 72\% & 0.157 \\
1.25 & 47\% & 0.097 & 74\% & 0.180 \\
1.50 & 29\% & 0.090 & 76\% & 0.195 \\
\bottomrule
\end{tabular}
\caption{
\textbf{Rotation-strength ablation on SD1.5.}
The complete contamination-aware correction (C5)
achieves higher target accuracy than subject-only
editing (C1) across all evaluated strengths.
}
\label{tab:token_strength_ablation}
\end{table}

\textbf{Results.}
As shown in Table~\ref{tab:token_strength_ablation},
the complete correction consistently achieves higher
target accuracy than subject-only editing across
all rotation strengths. Increasing the rotation
strength alone does not compensate for omitting
contaminated token positions: C1 reaches a maximum
accuracy of 53\%, whereas C5 achieves up to 76\%.
This confirms that the improvement arises from
correcting propagated concept information rather
than simply applying a stronger transformation.

Overall, these ablations demonstrate the importance
of contamination-aware token correction for effective
subject replacement. However, token propagation is
not the sole contributor to background drift.
The denoising intervention step provides an additional
mechanism for controlling scene preservation, as
investigated in Appendix~\ref{app:rebuttal_injection}.

\section{Geodesic Traversal}
\label{geo-traversal}

We perform semantic edits on the unit hypersphere
$\mathbb{S}^{D-1}=\{x\in\mathbb{R}^{D}:\|x\|_2=1\}$, where semantic similarity
is governed by angular distance rather than Euclidean magnitude.
Given normalized embeddings $u,v\in\mathbb{S}^{D-1}$, their geodesic distance is
$
\theta(u,v)=\arccos(u^\top v).
$
The shortest path between them is the great-circle geodesic:
\begin{equation}
\mathrm{Geo}(u,v;\lambda)
=
\frac{\sin((1-\lambda)\theta)}{\sin\theta}u
+
\frac{\sin(\lambda\theta)}{\sin\theta}v,
\qquad
\lambda\in[0,1].
\end{equation}
This corresponds to spherical interpolation (slerp), which preserves the
intrinsic geometry of the embedding manifold.

To perform local edits, we additionally use logarithmic and exponential maps
between the hypersphere and its tangent space.
The logarithmic map projects a point $v\in\mathbb{S}^{D-1}$ onto the tangent
space at $u$:
$$
\mathrm{Log}_{u}(v)
=
\frac{\theta}{\sin\theta}
\left(v-\cos\theta\,u\right),
\qquad
\theta=\arccos(u^\top v),
$$
producing a tangent vector pointing from $u$ toward $v$.
Conversely, the exponential map lifts a tangent vector
$\xi\in T_u\mathbb{S}^{D-1}$ back onto the hypersphere:
$$
\mathrm{Exp}_{u}(\xi)
=
\cos(\|\xi\|_2)u
+
\sin(\|\xi\|_2)
\frac{\xi}{\|\xi\|_2}.
$$
A geodesic step along tangent direction $d$ with strength $\lambda$ is thus
$
\mathrm{Exp}_{u}(\lambda d)
=
\cos(\lambda)u+\sin(\lambda)d.
$

HEART-S uses the geodesic operator to rotate a concept-aligned component from
source anchor $\mu_s$ toward target anchor $\mu_t$:
$
\mu_{s\rightarrow t}
=
\mathrm{Geo}(\mu_s,\mu_t;\lambda).
$
HEART-A instead computes an attribute direction in the tangent space and applies
an exponential-map update from the current token embedding:
$
\hat{h}^{(p)}
=
\mathrm{Exp}_{\tilde{h}^{(p)}}(\lambda d_a).
$

\section{Experimental Setup}
\label{app:experimental_setup}

\subsection{Setup Details}

\paragraph{Evaluation Metrics.}
We evaluate subject replacement using the following metrics:
(1) target-subject classification accuracy (\textbf{Acc}),
computed using zero-shot CLIP classification between source and
target concepts;
(2) image-text alignment (\textbf{CLIP-T}), measured as cosine
similarity between CLIP image and text embeddings;
(3) DINOv2 classification accuracy (\textbf{DINO Acc.}),
which provides an additional measure of target-concept acquisition;
and (4) source-concept persistence, measuring whether the original
subject remains after editing.

We additionally report perceptual similarity using
\textbf{LPIPS} (\cite{zhang2018unreasonable}).
However, full-image LPIPS measures both intended subject changes
and unintended background modifications. To distinguish these
effects, we use Grounded-SAM (\cite{ren2024grounded}) to obtain
subject masks and compute foreground LPIPS (\textbf{FG-LPIPS})
and background LPIPS (\textbf{BG-LPIPS}). Higher FG-LPIPS
indicates greater subject-region modification, whereas lower
BG-LPIPS indicates better background preservation.
We further measure background structural and semantic consistency
using masked SSIM (\cite{wang2004image}) and background DINOv2
similarity (\textbf{BG-DINO}) (\cite{oquab2023dinov2}), respectively.
For methods without explicit segmentation masks, the same
Grounded-SAM-based evaluation masks are used to ensure
consistent region-aware comparisons across methods.

For attribute control, we report $\Delta$CLIP to measure
attribute-related semantic changes and full-image LPIPS
to evaluate overall perceptual differences. We additionally
report FG-LPIPS and BG-LPIPS to distinguish intended
attribute modifications from unintended scene changes.

Following the evaluation protocol of Text Sliders
(\cite{chiu2026text}), our original evaluation generates
500 image pairs per attribute using the base prompt
\textit{``image of a person, photorealistic''} across seeds
$0\ldots499$, considering four attributes:
\textit{age}, \textit{smile}, \textit{curly hair}, and
\textit{chubby}. The extended evaluation considers
20 attributes spanning human and non-human concepts,
using 25 prompts and three random seeds per attribute.

\paragraph{Baselines.}
For subject replacement, we compare against
SEGA (\cite{brack2023sega}),
LEDITS++ (\cite{brack2024ledits++}),
and Direct Swap, which regenerates images directly from
the edited prompt without preservation constraints.
SEGA performs semantic guidance in the diffusion
noise-prediction space, whereas LEDITS++ relies on
latent inversion and attention-based editing.
We additionally consider \textbf{HEART (earlier injection)}
to examine the effect of denoising intervention timing
on subject replacement and background preservation.

For attribute control, we compare against
Concept Sliders (\cite{gandikota2024concept}),
Attribute Control (\cite{baumann2025continuous}), and Text Sliders (\cite{chiu2026text}).
We additionally include Prompt Rewrite, which directly
regenerates an image using the modified attribute prompt
without enforcing preservation of the original scene.

\paragraph{Implementation Details.}
All experiments are performed on NVIDIA H200 GPUs and A100 GPUs.
Subject replacement experiments are conducted primarily
on SD\,1.5 and SDXL, with additional evaluations on
SD3.5-medium and FLUX-Schnell. The extended benchmark
consists of 80 source-target concept pairs spanning
seven semantic categories, with three prompt templates
and three random seeds per pair.
For HEART-S, geodesic traversal is performed in
normalized text embedding space using edit strengths
$\lambda\in[-1,1]$ and prompt-dependent edit-start schedules (10-20\% of denoising steps).
We additionally evaluate different denoising intervention
steps to characterize the trade-off between subject
acquisition and background preservation, as detailed in
Appendix \ref{app:rebuttal_injection}.

For HEART-A, we use SDXL as the primary evaluation
backbone. In the original attribute-control evaluation,
images are generated using 30 diffusion steps at
resolution $512\times512$.
Attribute traversal is performed using geodesic updates
in tangent space, with separate controls for CLIP-L
and OpenCLIP-G embeddings.
We use the best-performing configuration:
geodesic mode,
\texttt{edit\_start\_step=2},
and edit strengths selected from uniformly sampled
traversal ranges.

The inference-time editing process, including concept-anchor
estimation from the prompt pool, takes approximately
37\,s for five generated images in the original setup.
For the extended 20-attribute benchmark, we additionally
report the average total computational cost per attribute,
including the attribute-specific setup, to enable comparison
with training-based baselines.

\subsection{License Information}
We will make our code publicly available on acceptance. We use the following models in our work:
\begin{itemize}
\item SD 1.5: https://huggingface.co/stable-diffusion-v1-5/stable-diffusion-v1-5
\item SDXL: https://huggingface.co/stabilityai/stable-diffusion-xl-base-1.0
\item SD3.5-medium: https://huggingface.co/stabilityai/stable-diffusion-3.5-medium
\item Flux-Schnell: https://huggingface.co/black-forest-labs/FLUX.1-schnell
\end{itemize}

\subsection{Prompts used for Subject Control}

\textbf{Evaluation Setup.}
For subject replacement, we follow a paired-prompt protocol
where the source concept is replaced with the target concept
while keeping the remaining scene description unchanged.
We evaluate 80 source--target concept pairs across seven
semantic categories: animals, natural entities, objects,
vehicles, food, clothing, and furniture. These include both
semantically related transitions (e.g., cat $\rightarrow$ dog)
and more distinct transformations (e.g., tree $\rightarrow$
mountain). Each concept pair is evaluated using three prompt
templates and three random seeds, resulting in 720 paired
evaluations per backbone.

The same evaluation protocol is used across SD1.5, SDXL,
SD3-medium, and FLUX-Schnell to assess subject replacement
and background preservation across different text encoders.
For each source-target pair, the source and target prompts
differ only in the subject concept, allowing the evaluation
to measure the intended subject modification while keeping
the surrounding textual context fixed.
Representative prompt pairs are provided in
Table~\ref{tab:subject_prompts}.

\subsection{Prompts used for Attribute Control}

\textbf{Evaluation Setup.}
For attribute control, we follow the paired semantic-direction
protocol of Text Sliders (\cite{chiu2026text}), where positive
and negative prompt pairs define the source and target
attribute anchors for geodesic traversal.
We evaluate 20 attributes spanning global, regional, and
surface-level modifications, including both human and
non-human concepts. Approximately half of the attributes
involve human appearance (e.g., age, smile, hair texture,
body shape, and makeup), while the remaining attributes
involve non-human properties (e.g., car age and bed size).

Each attribute is evaluated using 25 prompts and three
random seeds, yielding 75 paired observations per attribute
and 1,500 observations across the benchmark.
The source and target prompts describe opposite attribute
states while retaining the underlying subject and context.
Multiple prompt pairs can be used to estimate the
attribute-conditioned concept anchors, reducing dependence
on any individual phrasing. Representative positive and
negative prompt pairs are provided in
Table~\ref{tab:attribute_prompts}.

\begin{table*}[t]
\centering
\caption{Prompt pairs used for subject replacement evaluation.}
\label{tab:subject_prompts}
\small
\setlength{\tabcolsep}{6pt}
\renewcommand{\arraystretch}{1.05}

\begin{tabular}{p{3.2cm}p{5.8cm}p{5.8cm}}
\toprule

\textbf{Edit Type}
&
\textbf{Source Prompt}
&
\textbf{Edited Prompt}
\\

\midrule

Car $\rightarrow$ Truck
&
``A car on a road''
&
``A truck on a road''
\\

Dog $\rightarrow$ Wolf
&
``A dog in a forest''
&
``A wolf in a forest''
\\

Cat $\rightarrow$ Dog
&
``A photo of a cat sitting on a chair''
&
``A photo of a dog sitting on a chair''
\\

Apple $\rightarrow$ Orange
&
``An apple on a branch''
&
``An orange on a branch''
\\

Horse $\rightarrow$ Zebra
&
``A horse running in a field''
&
``A zebra running in a field''
\\

Bird $\rightarrow$ Eagle
&
``A bird flying over mountains''
&
``An eagle flying over mountains''
\\

Boat $\rightarrow$ Ship
&
``A boat floating on the ocean''
&
``A ship floating on the ocean''
\\

Bicycle $\rightarrow$ Motorcycle
&
``A bicycle parked beside a building''
&
``A motorcycle parked beside a building''
\\

\bottomrule
\end{tabular}
\end{table*}

\begin{table*}[t]
\centering
\caption{Prompt pairs used for attribute control evaluation.}
\label{tab:attribute_prompts}
\small
\setlength{\tabcolsep}{6pt}
\renewcommand{\arraystretch}{1.05}

\begin{tabular}{p{3.0cm}p{5.8cm}p{5.8cm}}
\toprule

\textbf{Attribute}
&
\textbf{Negative / Source Prompt}
&
\textbf{Positive / Target Prompt}
\\

\midrule

Age
&
``A young person, photorealistic''
&
``An old person, photorealistic''
\\

Smile
&
``A serious person, photorealistic''
&
``A smiling person, photorealistic''
\\

Hair Texture
&
``A person with straight hair, photorealistic''
&
``A person with curly hair, photorealistic''
\\

Body Shape
&
``A slim person, photorealistic''
&
``A chubby person, photorealistic''
\\

Beard
&
``A clean-shaven man, photorealistic''
&
``A bearded man, photorealistic''
\\

Glasses
&
``A person without glasses, photorealistic''
&
``A person wearing glasses, photorealistic''
\\

Hair Color
&
``A person with black hair, photorealistic''
&
``A person with blonde hair, photorealistic''
\\

Expression
&
``A calm person, photorealistic''
&
``An angry person, photorealistic''
\\

\bottomrule
\end{tabular}
\end{table*}

\section{Additional Quantitative Results and Ablations}
\label{app:rebuttal_results}

\subsection{Denoising Intervention Timing}
\label{app:rebuttal_injection}

\textbf{Setup.}
We evaluate the effect of denoising intervention timing
on subject replacement and background preservation
across SD1.5, SDXL, SD3, and FLUX-Schnell using
80 concept pairs and three random seeds.
We vary the injection step while keeping other
generation settings fixed, and measure target
accuracy, FG-LPIPS, and BG-LPIPS.

\begin{table}[!htbp]
\centering
\small
\setlength{\tabcolsep}{5pt}
\renewcommand{\arraystretch}{1.12}

\textbf{(a) Selected operating points}

\vspace{4pt}
\begin{tabular}{lccc}
\toprule
Model & Immediate & Selected & Late \\
\midrule
SDXL (50)  & 90/.339/.618 & 77/.228/.491 & 58/.152/.366 \\
SD1.5 (50) & 91/.310/.616 & 87/.241/.544 & 76/.167/.423 \\
SD3 (28)   & 93/.330/.648 & 89/.171/.463 & 60/.064/.274 \\
FLUX-S (4) & 57/.262/.522 & 2/.008/.038  & -- \\
\bottomrule
\end{tabular}

\vspace{12pt}

\textbf{(b) Full injection-step sweep}

\vspace{4pt}
\begin{tabular}{ccc}
\toprule
Injection (\%) & SDXL & SD1.5 \\
\midrule
0  & 90/.339 & 91/.310 \\
2  & 91/.319 & 91/.304 \\
5  & 88/.295 & 91/.293 \\
10 & 85/.270 & 93/.279 \\
15 & 79/.246 & 88/.261 \\
\textbf{20} & \textbf{77/.228} & \textbf{87/.241} \\
25 & 58/.152 & 76/.167 \\
30 & 46/.103 & 62/.104 \\
40 & 29/.067 & 44/.061 \\
50 & 16/.045 & 29/.036 \\
60 & 9/.030  & 19/.019 \\
80 & 2/.012  & 8/.004 \\
\bottomrule
\end{tabular}

\vspace{4pt}
\caption{
\textbf{Denoising intervention timing.}
(a) Target Acc. (\%)/BG-LPIPS/FG-LPIPS at different
injection points. Late denotes approximately twice
the selected delay.
(b) Complete injection sweep reporting
Target Acc. (\%)/BG-LPIPS.
Bold indicates the selected 20\% operating point.
}
\label{tab:injection_ablation}
\end{table}

\textbf{Results.}
As shown in Table~\ref{tab:injection_ablation},
delaying intervention reduces background drift but
progressively weakens subject modification.
For SD1.5 and SDXL, target accuracy drops sharply
beyond 20\%, motivating the selected operating point.
SD3 exhibits a similar trade-off, while FLUX-Schnell
offers limited flexibility due to its four-step
denoising trajectory.

\subsection{Robustness of Concept Anchor Estimation}
\label{app:rebuttal_anchor}
\begin{wraptable}{r}{0.46\columnwidth}
\vspace{-14pt}
\centering
\scriptsize
\setlength{\tabcolsep}{3pt}
\renewcommand{\arraystretch}{1.0}

\begin{tabular}{lc}
\toprule
\textbf{Setting} & \textbf{Deviation} \\
\midrule
\multicolumn{2}{l}{\textit{Prompt-pool size}} \\
$K=5$  & $7^\circ$-$18^\circ$ \\
$K=40$ & $2^\circ$-$5^\circ$ \\
\midrule
\multicolumn{2}{l}{\textit{Template diversity}} \\
Across templates & $11^\circ$-$14^\circ$ \\
Across concepts & $48^\circ$-$66^\circ$ \\
\bottomrule
\end{tabular}

\caption{\textbf{Anchor robustness.}
Angular variation across prompt pools and template families.}
\label{tab:anchor_robustness}
\vspace{-16pt}
\end{wraptable}
\textbf{Setup.}
We evaluate the sensitivity of concept anchor estimation to
prompt-pool size and diversity across SD1.5, SDXL, SD3, and
FLUX. First, we vary the number of prompts per concept,
$K\in\{5,10,20,40,80\}$, and measure the bootstrap angular
deviation from a reference anchor. Second, we compare anchors
estimated from four disjoint template families: photographic,
stylistic, contextual, and automatically generated prompts.

\textbf{Results.}
As shown in Table~\ref{tab:anchor_robustness},
angular deviation decreases from $7^\circ$-$18^\circ$
with five prompts to $2^\circ$-$5^\circ$ with 40 prompts,
indicating stable anchor estimation with moderate prompt
pools. Across template families, anchor variation remains
within $11^\circ$-$14^\circ$, substantially smaller than
the inter-concept separation of $48^\circ$-$66^\circ$.
These results demonstrate that concept anchors are robust
to prompt-pool size and diversity, with concept-specific
differences dominating template-induced variation.

\subsection{Cross-Architecture Subject Control}
\label{app:rebuttal_subject}

\textbf{Setup.}
We evaluate HEART-S across four diffusion backbones:
SD1.5, SDXL, SD3-medium, and FLUX-Schnell, using
80 source--target concept pairs, three prompts, and
three random seeds per pair. These architectures
incorporate different text encoders, ranging from
CLIP-based conditioning to joint CLIP and T5
representations. We report target accuracy, CLIP-T,
background preservation metrics, and runtime.

\begin{table*}[t]
\centering
\scriptsize
\setlength{\tabcolsep}{4pt}
\renewcommand{\arraystretch}{1.05}

\resizebox{\textwidth}{!}{%
\begin{tabular}{llcccccc}
\toprule
\textbf{Backbone}
& \textbf{Edited Text Encoders}
& \textbf{Target Acc.} $\uparrow$
& \textbf{CLIP-T} $\uparrow$
& \textbf{BG-LPIPS} $\downarrow$
& \textbf{Masked SSIM} $\uparrow$
& \textbf{BG-DINO} $\uparrow$
& \textbf{Runtime (s)} $\downarrow$ \\
\midrule
SD1.5
& CLIP-L
& 0.496 & 0.260 & 0.197 & 0.757 & 0.837 & 2.10 \\

SDXL
& CLIP-L + OpenCLIP-G
& 0.311 & 0.248 & \textbf{0.137}
& \textbf{0.883} & \textbf{0.885} & 2.98 \\

SD3-medium
& CLIP-L + CLIP-G + T5
& \textbf{0.685} & 0.258 & 0.307
& 0.757 & 0.777 & 2.25 \\

FLUX-Schnell
& T5
& 0.622 & \textbf{0.264} & 0.460
& 0.514 & 0.721 & \textbf{0.99} \\
\bottomrule
\end{tabular}%
}

\caption{
\textbf{Cross-architecture subject control with HEART-S.}
Evaluation across diffusion backbones with different text
conditioning architectures. Bold indicates the best value
in each column.
}
\label{tab:cross_arch_subject}
\end{table*}

\textbf{Results.}
Table~\ref{tab:cross_arch_subject} demonstrates that
HEART-S enables subject replacement across different
diffusion architectures without finetuning or inversion.
SDXL achieves the best background preservation, whereas
SD3-medium and FLUX-Schnell obtain higher target
accuracy. However, FLUX-Schnell exhibits greater
background drift, consistent with its limited four-step
denoising trajectory. These results demonstrate the
applicability of geometric subject control across
heterogeneous text-conditioning architectures, with
model-dependent editing--preservation trade-offs.

\subsection{Cross-Architecture Attribute Control}
\label{app:rebuttal_attributes}

\textbf{Setup.}
We evaluate HEART-A across SD1.5, SDXL, SD3-medium,
and FLUX-Schnell using twenty attributes: human
attributes (e.g. age, weight, muscularity, smile, curly hair,
long hair, surprise, makeup etc.) and non-human
attributes (car age, bed size, plant height etc.). Each attribute
uses 25 prompts and three random seeds, evaluated
at each backbone's tuned operating point. We compare against Prompt Rewrite, which directly generates images using attribute-modified prompts without preserving the original scene.

\begin{table*}[t]
\centering
\scriptsize
\setlength{\tabcolsep}{3pt}
\renewcommand{\arraystretch}{1.05}

\resizebox{\textwidth}{!}{%
\begin{tabular}{llcccccccc}
\toprule
\textbf{Backbone}
& \textbf{Edited Text Encoders}
& $\boldsymbol{\Delta}$\textbf{CLIP} $\uparrow$
& \textbf{Succ.} $\uparrow$
& \textbf{CLIP-T} $\uparrow$
& \textbf{FG-LPIPS} $\uparrow$
& \textbf{BG-LPIPS} $\downarrow$
& \textbf{FG/BG} $\uparrow$
& \textbf{Masked SSIM} $\uparrow$
& \textbf{BG-DINO} $\uparrow$ \\
\midrule

\multicolumn{10}{l}{\textit{HEART-A (training-free, no inversion)}} \\
SD1.5 & CLIP-L
& +0.0086 & 0.423 & 0.267 & 0.090
& 0.072 & 1.2 & 0.830 & 0.937 \\

SDXL & CLIP-L + OpenCLIP-G
& +0.0152 & 0.820 & 0.261 & 0.186
& 0.197 & 0.9 & 0.707 & 0.845 \\

SD3-medium & CLIP-L + CLIP-G + T5
& +0.0164 & 0.756 & 0.248 & 0.207
& 0.165 & 1.2 & 0.649 & 0.879 \\

FLUX-Schnell & T5
& +0.0083 & 0.665 & 0.256 & 0.295
& 0.288 & 1.0 & 0.439 & 0.847 \\

\midrule
\multicolumn{10}{l}{\textit{Prompt Rewrite (reference baseline)}} \\

SD1.5 & --
& +0.0432 & 0.928 & 0.283 & 0.355
& 0.430 & 0.8 & 0.365 & 0.583 \\

SDXL & --
& +0.0508 & 0.969 & 0.284 & 0.347
& 0.454 & 0.8 & 0.418 & 0.587 \\

SD3-medium & --
& +0.0568 & 0.987 & 0.277 & 0.413
& 0.475 & 0.9 & 0.285 & 0.631 \\

FLUX-Schnell & --
& +0.0519 & 0.991 & 0.279 & 0.421
& 0.497 & 0.9 & 0.240 & 0.598 \\

\bottomrule
\end{tabular}%
}

\caption{
\textbf{Cross-architecture attribute control with HEART-A.}
Twenty attributes evaluated using 25 prompts and three seeds
per attribute. Succ. denotes the fraction of edits with
positive $\Delta$CLIP, while FG/BG measures the ratio
of foreground to background modification.
Prompt Rewrite serves as a regeneration baseline.
Metrics are computed after resizing images to
$512\times512$; preservation metrics should be
compared within each backbone.
}
\label{tab:t4_attr}
\end{table*}

\textbf{Results.}
Table~\ref{tab:t4_attr} shows that HEART-A enables
attribute manipulation across all four diffusion
backbones. SDXL achieves the highest success rate
(0.820), while SD3-medium obtains the largest average
$\Delta$CLIP (+0.0164). Although Prompt Rewrite achieves
stronger semantic changes, these are accompanied by
substantially greater background modifications.
HEART-A consistently achieves lower BG-LPIPS and
higher masked SSIM and BG-DINO across all backbones,
demonstrating improved scene preservation without
training or inversion.

\subsection{Additional Attribute-Level Results}
\label{app:rebuttal_attribute_details}

\textbf{Setup.}
To further examine background preservation across
different attribute types, we compare HEART-A with
Concept Sliders, Text Slider, Attribute Control, and
Prompt Rewrite on five representative attributes:
surprised, makeup, long hair, car age, and bed size.
We report BG-LPIPS at matched attribute-editing
strength ($\Delta$CLIP) on SDXL, allowing direct
comparison of background preservation.

\begin{table}[t]
\centering
\scriptsize
\setlength{\tabcolsep}{3pt}
\renewcommand{\arraystretch}{1.0}
\begin{tabular}{@{}lccccc@{}}
\toprule
\textbf{Attribute}
& \textbf{HEART-A}
& \shortstack{\textbf{Concept}\\\textbf{Sliders}}
& \shortstack{\textbf{Text}\\\textbf{Slider}}
& \shortstack{\textbf{Attribute}\\\textbf{Control}}
& \shortstack{\textbf{Prompt}\\\textbf{Rewrite}} \\
\midrule
Surprised & \textbf{.017} & .196 & .023 & .047 & .336 \\
Makeup    & \textbf{.016} & .459 & .028 & .050 & .396 \\
Long Hair & .040 & .164 & .039 & \textbf{.030} & .453 \\
Car Age   & \textbf{.049} & .340 & .064 & .112 & .452 \\
Bed Size  & \textbf{.043} & .237 & .058 & .063 & .374 \\
\bottomrule
\end{tabular}
\caption{\textbf{Attribute-level BG-LPIPS} ($\downarrow$)
at matched $\Delta$CLIP on SDXL. Bold indicates best.}
\label{tab:attribute_matched_strength}
\end{table}

\textbf{Results.}
Table~\ref{tab:attribute_matched_strength} shows that
HEART-A achieves the lowest background drift on four
of the five representative attributes, covering both
human and non-human edits. For long-hair editing,
Attribute Control achieves slightly better preservation
(0.030 versus 0.040). These results further support
the background-preservation advantages observed in
the broader attribute-control evaluation, while
demonstrating that the gains are not uniform across
all attributes.

\section{Limitations of our work}
HEART operates purely in text embedding space and does not directly manipulate
cross-attention maps, latent trajectories, or diffusion features. As a result, extremely fine-grained spatial edits or precise object-localized
modifications can remain challenging, particularly in highly compositional scenes with strong semantic overlap between subjects. Similarly, while our inversion-free formulation enables efficient and general
editing, inversion-based pipelines can provide stronger reconstruction fidelity
for certain real-image editing tasks.
Also, models like FLUX-Schnell which have very few denosing steps might not be able to showcase a lot of control ability. Our method depends upon the denoising steps to be at least more than 5.

\section{SD3.5-medium and FLUX visual results}
\begin{figure*}[t]
\centering
\includegraphics[width=\textwidth]{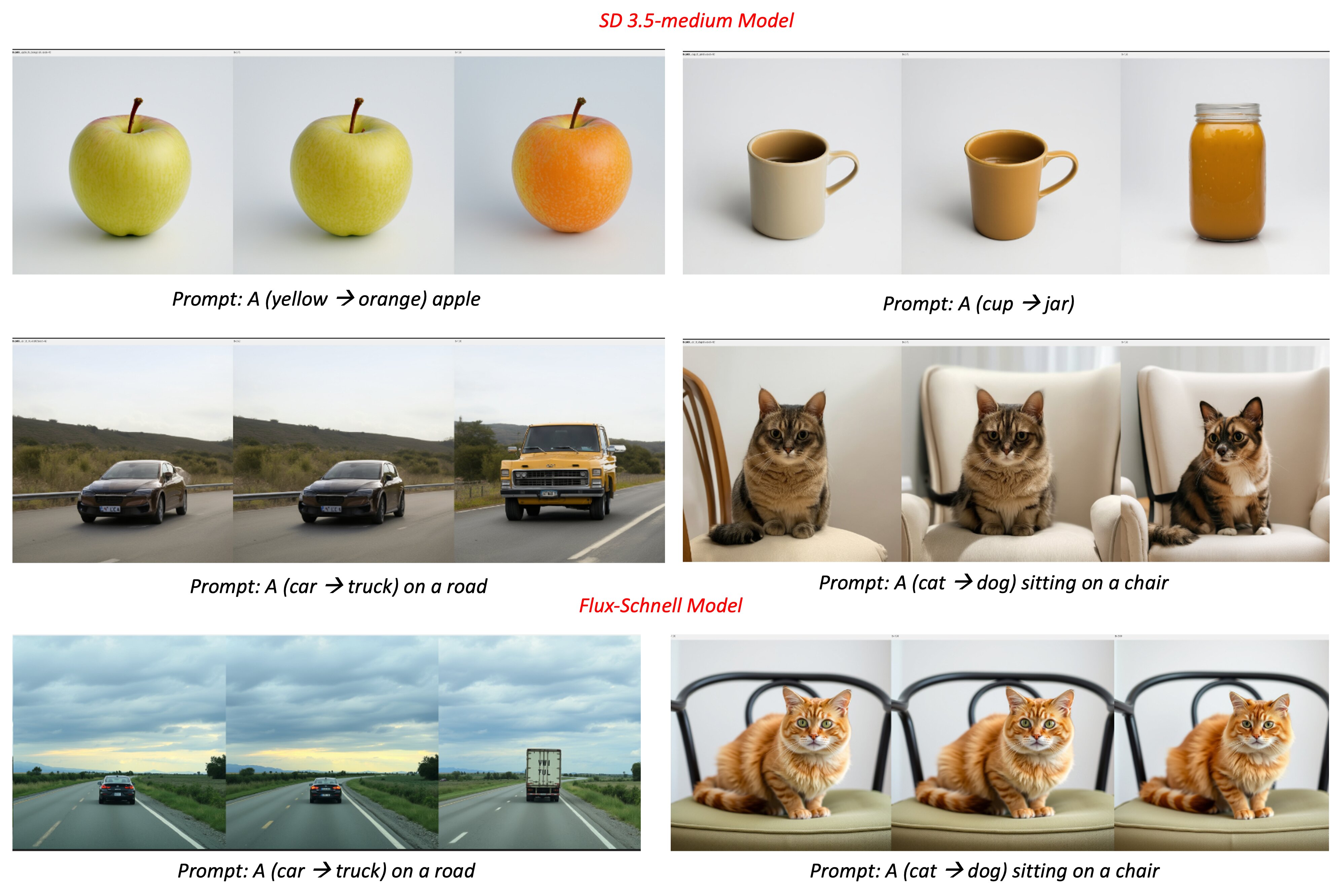}

\caption{
Cross-architecture subject control using HEART-S.
Qualitative subject replacement results on SD\,3.5-Medium (top two rows)
and FLUX-Schnell (bottom row).
HEART performs semantic traversal directly in text embedding space and
generalizes across both CLIP- and T5-conditioned diffusion architectures. Despite large subject changes (e.g., apple$\rightarrow$orange,
car$\rightarrow$truck, and cat$\rightarrow$dog), the method preserves
scene layout, viewpoint, lighting, and contextual structure while modifying
only the target concept. Note, due to limited denoising steps (4) for Flux-Schnell model, the control acts on few concepts only.
}
\label{fig:cross_arch_subject_control}

\end{figure*}

\begin{figure*}[t]
\centering
\includegraphics[width=\textwidth]{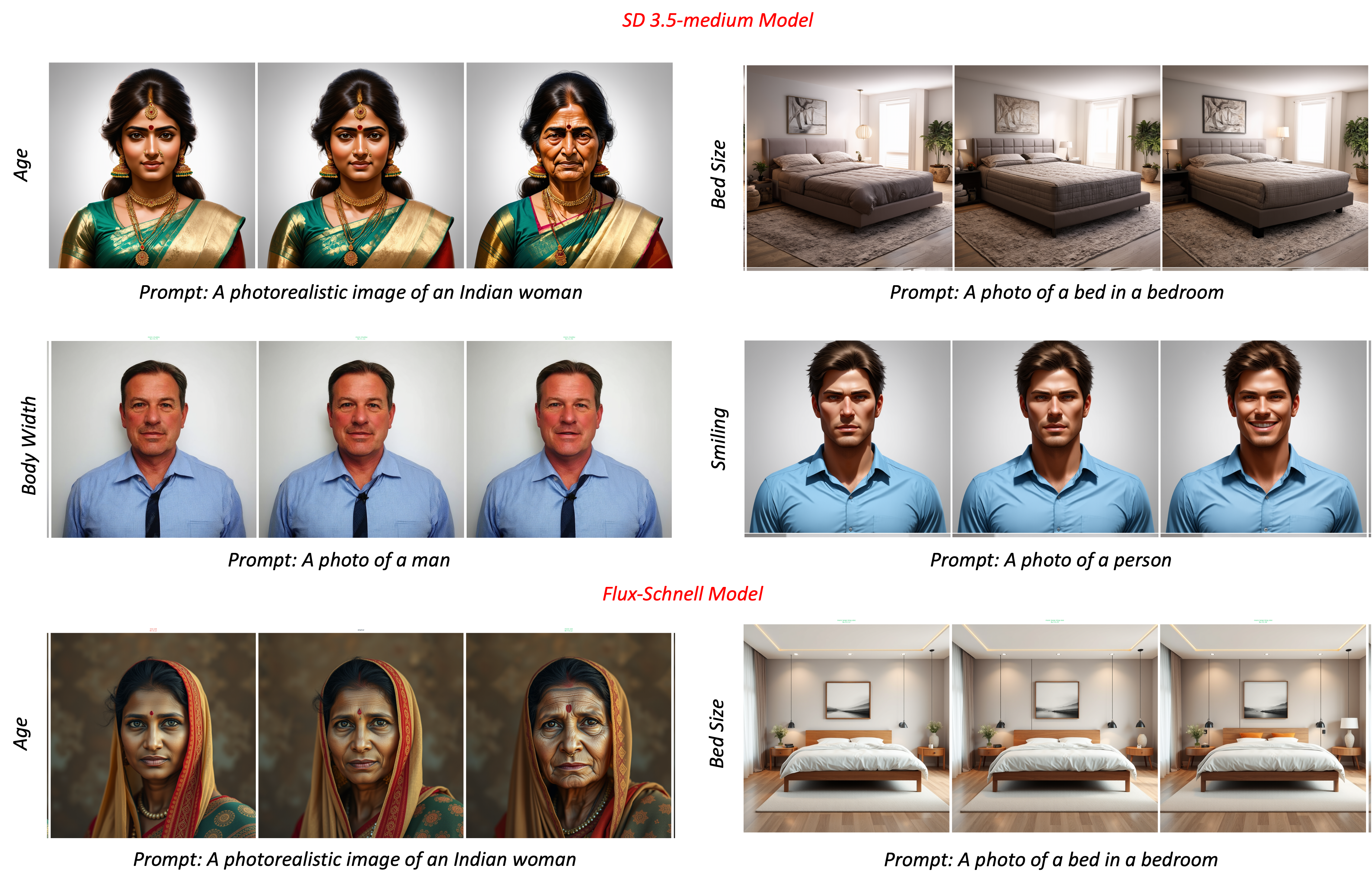}

\caption{
\textbf{Cross-architecture attribute control using HEART-A.}
Qualitative attribute modification results on SD\,3.5-Medium
(top two rows) and FLUX-Schnell (bottom row).
HEART performs geodesic attribute traversal within the Kent-concept distribution in text embedding space and generalizes across both CLIP- and T5-conditioned diffusion architectures. The examples demonstrate progressive changes in age (top-left and bottom-left), bed size (top-right and bottom-right), and facial expressions (middle row), while largely preserving subject identity, clothing, scene layout, and background details. Note that FLUX-Schnell, with only four denoising steps, offers limited flexibility for fine-grained attribute control.
}
\label{fig:cross_arch_attribute_control}

\end{figure*}

We observe consistent semantic control on SD\,3.5-Medium across diverse
subject replacement tasks, demonstrating that HEART generalizes effectively to
modern multi-encoder diffusion architectures in Figure ~\ref{fig:cross_arch_subject_control}. On FLUX-Schnell, semantic edits
remain observable, particularly for large object-level transitions such as
car$\rightarrow$truck. However, because FLUX-Schnell operates with an extremely
short denoising trajectory (only 4 diffusion steps), the influence of text-space
geometric traversal is comparatively limited. As a result, subtle or fine-grained
subject changes, such as cat$\rightarrow$dog, may not fully propagate through
the denoising process, leading to weaker visible semantic transitions.
Nevertheless, the results indicate that directional geometric control persists
even under highly compressed diffusion schedules.

We further see consistent attribute control on SD\,3.5-Medium
across both human and non-human attributes, as shown in
Figure~\ref{fig:cross_arch_attribute_control}. HEART-A enables
progressive modifications in age, facial expressions, and bed size
while largely preserving subject identity, clothing, and surrounding
scene structure. These results demonstrate that geodesic attribute
traversal generalizes to modern multi-encoder diffusion architectures.
On FLUX-Schnell, attribute changes such as aging modification remain observable. However, its short denoising trajectory of only four steps limits the degree of fine-grained
control, particularly for subtle attribute variations.
Overall, the results show that attribute-level geometric
control remains feasible across different text-conditioning
architectures.



\end{document}